\documentclass[runningheads]{llncs}
\usepackage[T1]{fontenc}
\usepackage{graphicx}
\usepackage{amsmath}
\usepackage{subcaption}
\usepackage{booktabs}
\usepackage{colortbl}
\usepackage{multicol}
\usepackage{multirow}
\usepackage{xspace}
\usepackage{hyperref}

\usepackage[inline]{enumitem}
\usepackage[backend=biber,
bibstyle=ieee,
citestyle=numeric,
sorting=nyt,
maxbibnames=10]
{biblatex}

\usepackage{glossaries}

\newacronym[nonumberlist]{rats}{RATS}{Rapid Augmentations for Time Series}
\newacronym[nonumberlist]{dct}{DCT}{Discrete Cosine Transform}
\newacronym[nonumberlist]{fft}{FFT}{Fast Fourier Transform}
\newacronym[nonumberlist]{app}{APP}{Amplitude and Phase Perturbation}
\newacronym[nonumberlist]{dtw}{DTW}{Dynamic Time Wraping}
\newacronym[nonumberlist]{dft}{DFT}{Discrete Fourier Transform}

\begin{document}

\glsdisablehyper

\title{Rapid Augmentations for Time Series (RATS):\newline{}
A High-Performance Library for\newline{}Time Series Augmentation}
\titlerunning{Rapid Augmentations for Time Series (RATS)}
\author{
Wadie Skaf$\,^{\dagger}$\orcidID{\href{https://orcid.org/0000-0002-4298-6694}{\includegraphics[scale=0.5]{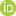}}} 
\and
Felix Kern$^{\dagger}$
\and 
Aryamaan Basu Roy$^{\dagger}$
\and
Tejas Pradhan$^{\dagger}$
\and \\
Roman Kalkreuth\orcidID{\href{https://orcid.org/0000-0003-1449-5131}{\includegraphics[scale=0.5]{orcid.png}}} \and
Holger Hoos\orcidID{\href{https://orcid.org/0000-0003-0629-0099}{\includegraphics[scale=0.5]{orcid.png}}}}

\authorrunning{W. Skaf et al.}
\institute{Chair for Artificial Intelligence Methodology (AIM), Faculty of Computer Science, RWTH Aachen University, Aachen, Germany \\ \vspace{0.8em}
%\email{\{skaf\}@uni-heidelberg.de}
$^{\dagger}$These authors contributed equally to this work.
}

\maketitle              % typeset the header of the contribution
\begin{abstract}
Time series augmentation is critical for training robust deep learning models, particularly in domains where labelled data is scarce and expensive to obtain.
% \hh{minor rewording (was awkward):}
However, existing augmentation libraries for time series, mainly written in Python, suffer from performance bottlenecks, where running time 
% \hh{please note: running time = the time something runs for; runtime is something different (as in runtime environment); please check and fix everywhere.}\ws{Done}
grows exponentially as dataset sizes increase---an aspect limiting their applicability in large-scale, production-grade systems.
We introduce \emph{\textbf{RATS}} (Rapid Augmentations for Time Series), a high-performance library for time series augmentation written in Rust with Python bindings (\emph{RATSpy}).
\emph{RATS} implements multiple augmentation methods spanning basic transformations, frequency-domain operations and time warping techniques, all accessible through a unified pipeline interface with built-in parallelisation.
Comprehensive benchmarking of \emph{RATSpy} versus a commonly used library (\emph{tasug}) on 143 datasets demonstrates that \emph{RATSpy} achieves an average speedup of 74.5\% over \emph{tsaug} (up to 94.8\% on large datasets), with up to 47.9\% less peak memory usage.

\end{abstract}
\section{Introduction}\label{introduction}

Time series analysis is a prevalent topic with multiple use cases, such as forecasting \parencite{de_2006_25_years_of_forecasting}, 
clustering \parencite{ma_2019_learning_ts_clustering, ma_2021_learning_ts_incomplete_clustering}, classification \parencite{fawaz_2019_deep_learning_ts_survey} 
and anomaly detection \parencite{skaf_2022_denoising}, in addition to applications in numerous sectors, including environmental science \parencite{chen_2018_wind_forecasting} and healthcare \parencite{skaf_2023_chickenpox_hun}.

Deep learning methods trained in a fully supervised manner have achieved strong results in time series classification \parencite{fawaz_2019_deep_learning_ts_survey, foumani_2024_dl_ts_survey, MoeEtAl26}; however, these results hinge on the availability of large quantities of labelled data that are typically expensive and 
% \hh{added:}
often infeasible to obtain.

Data augmentation has emerged as a feasible solution to artificially increase the amount of high-quality, expensive-to-obtain labelled data without incurring the high cost of collecting and labelling new data; moreover,it has been extensively used in contrastive-based 
% \hh{shouldn't that just be `contrastive' or `contrastive learning'?}\ws{We had a similar discussion while writing the TSRC paper and resorted to ``contrastive-based'' in the end.}
methods to generate the sample triplets (or pairs) required for contrastive loss when training models \parencite{meng_2023_unsupervised_ts_rl_review, zhang_2024_self_supervised_ts_survey, middlehurst_2024_bake-off-redux-ts, skaf_2025_tsrc, skaf_autotsaugment} and foundation models \parencite{yeh_2023_timeclr, sun_2024_test, ye_2024_tsfm_survey, skaf_autotsaugment}.
However, using data augmentation entails the following challenges:
\begin{enumerate*}[label=(\roman*), itemjoin={{, }}, itemjoin*={{, and }}]
    \item the computational cost of obtaining the augmented data
    \item the selection of suitable augmentation methods for each datasets and downstream task (it has been demonstrated that choosing an augmentation strategy is a critical step that can have larger impact than the network architecture  \parencite{lemley_2017_smartaugment, goodfellow_2016_deep_learning})
\end{enumerate*}.
Despite its importance, fewer tools have been developed to address the challenges 
% \hh{slightly reworded:}
of augmenting time series data, compared to data from other domains, such as computer vision, where many tools are available.
The most notable library for time series augmentation is \emph{tsaug} \parencite{tsaug}.
As a native Python library, \emph{tsaug} suffers from slow performance limited by the Python interpreter, hindering 
% \hh{try to avoid this repetition of `limited'}\ws{I replaced it with ``hindering''}\hh{well done!}
its application in larger-scale and performance-critical applications.

In this report, we introduce \gls{rats}, a high-performance library for time series augmentation.
\emph{\gls{rats}} is written in Rust, a compiled low-level language, and includes Python bindings to 
facilitate 
% \hh{modified:}
its use within existing pipelines.

In the following sections, we present the details of our high-performance time series augmentation library \emph{\gls{rats}} and its Python wrapper \emph{RATSpy}, highlighting the main features in Section \ref{sec:features}, followed by implementation details in Section \ref{sec:implementation}. Furthermore, we present benchmarking results and compare the performance of \emph{RATS} against \emph{tsaug} in Section \ref{sec:perf_eval}.

\section{Features}\label{sec:features}

\emph{\gls{rats}} and \emph{RATSpy} support a wide spectrum of augmentation techniques for univariate time series data.
In total, the libraries support 17 different augmenters, most of which are basic transformations, with some having more complex functionality.
% \hh{slightly modified -- check (until end of paragraph):}\ws{Done}
For each augmenter, it is possible to specify a probability with which it is executed for a given series in a batch.
This allows for the probabilistic application of augmenters that can increase the variability of series in a dataset.
These libraries also include the pipelining of augmenters.
Using pipelining, it is possible to compose multiple augmenters and apply them one after the other.
% \hh{one after the other? surely, the order can matter (depending on the augmenters in question) ...}\ws{Yes, it one after the other. Modified}
The methods implemented in the libraries can be classified into the following: basic augmentations, frequency domain transformations, frequency domain augmentations, time warping augmentations, and similarity measures (for assessing the quality of generated augmentations).
% \hh{add a sentence to explain why those last ones are needed.}\ws{Done. I added it between parenthesis.}

For simplicity, in the following, we use \emph{\gls{rats}} to collectively refer to both \emph{\gls{rats}} and \emph{RATSpy}.

\subsection{Basic Augmentations}\label{sec:basic_augmentations}

\begin{figure*}[t]
\centering
\includegraphics[width=0.4\textwidth]{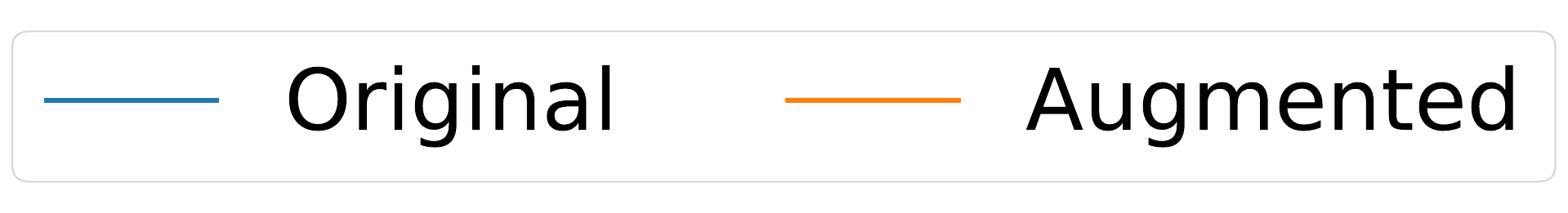}\\[2mm]
\includegraphics[width=0.3\textwidth]{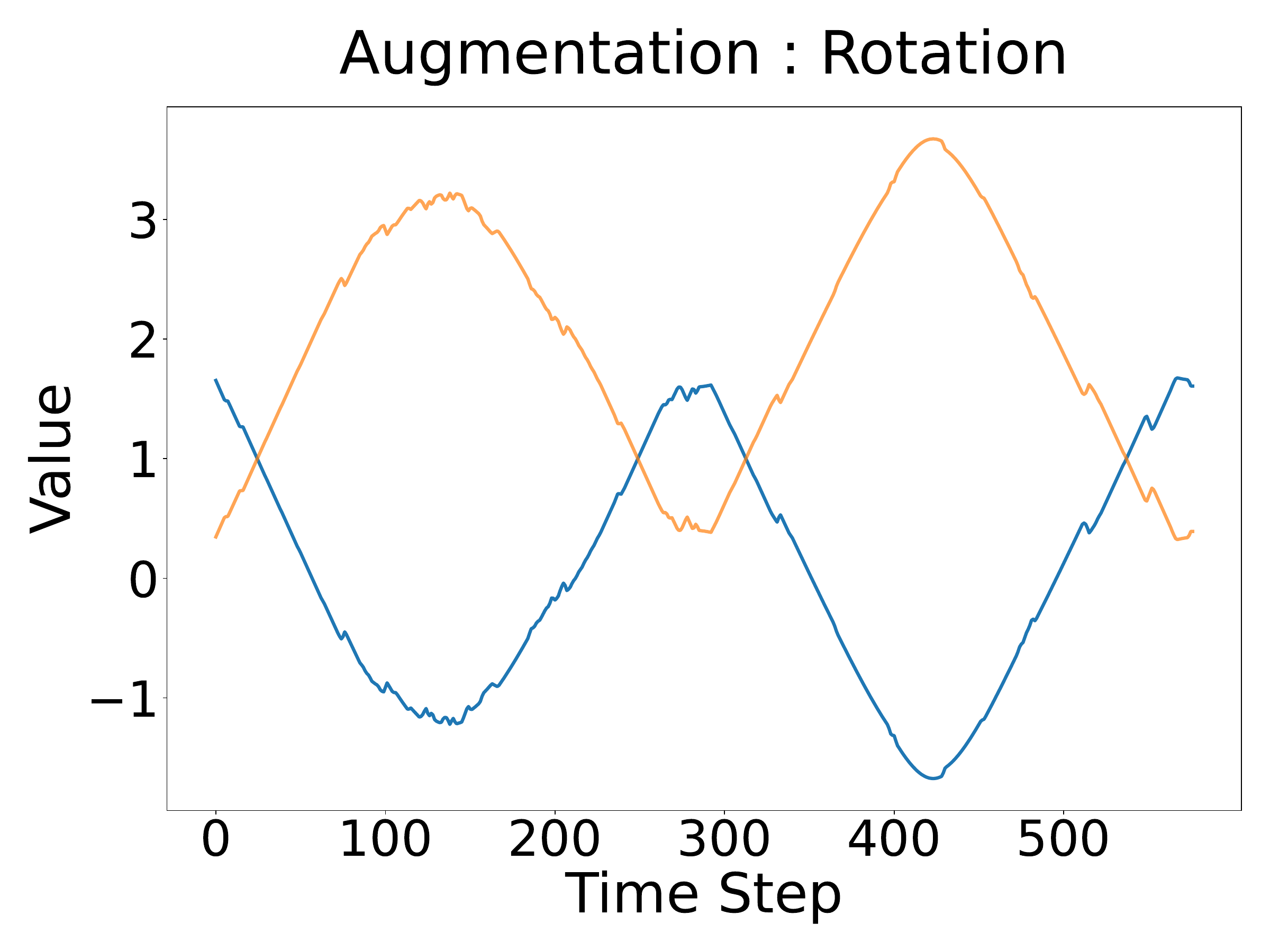}\hspace{3mm}
\includegraphics[width=0.3\textwidth]{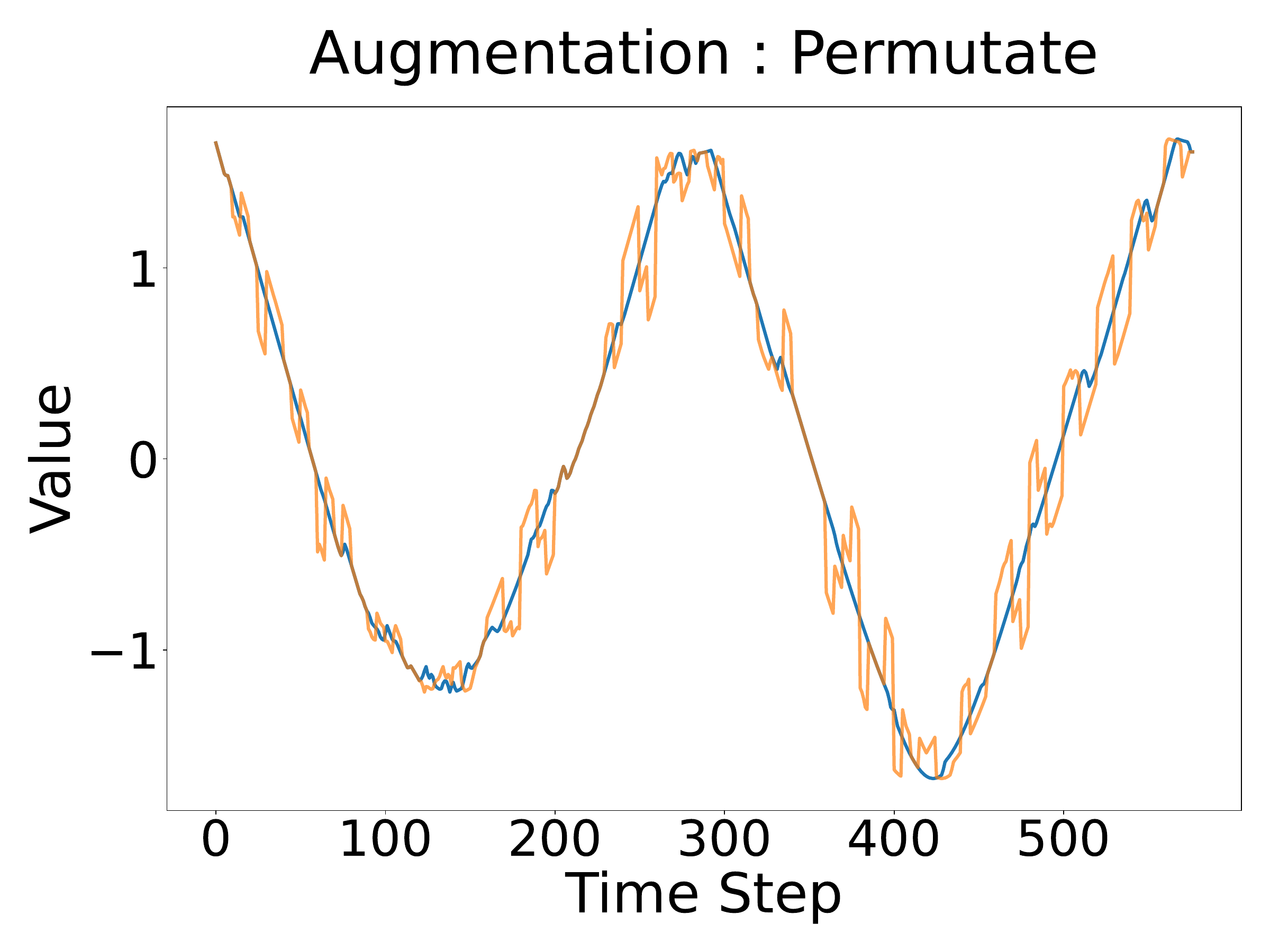}\hspace{3mm} 
\includegraphics[width=0.3\textwidth]{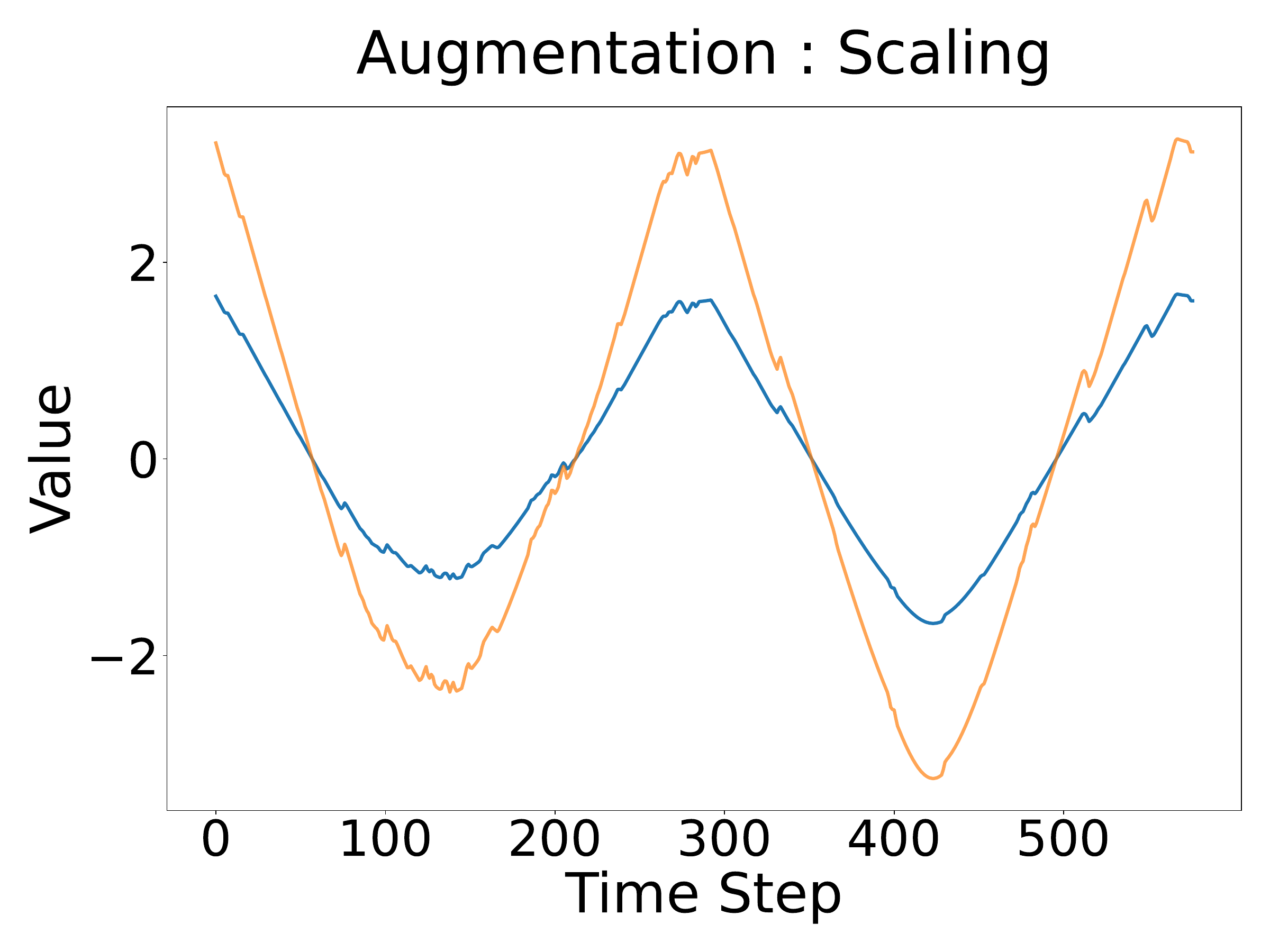}\\[4mm]
\includegraphics[width=0.3\textwidth]{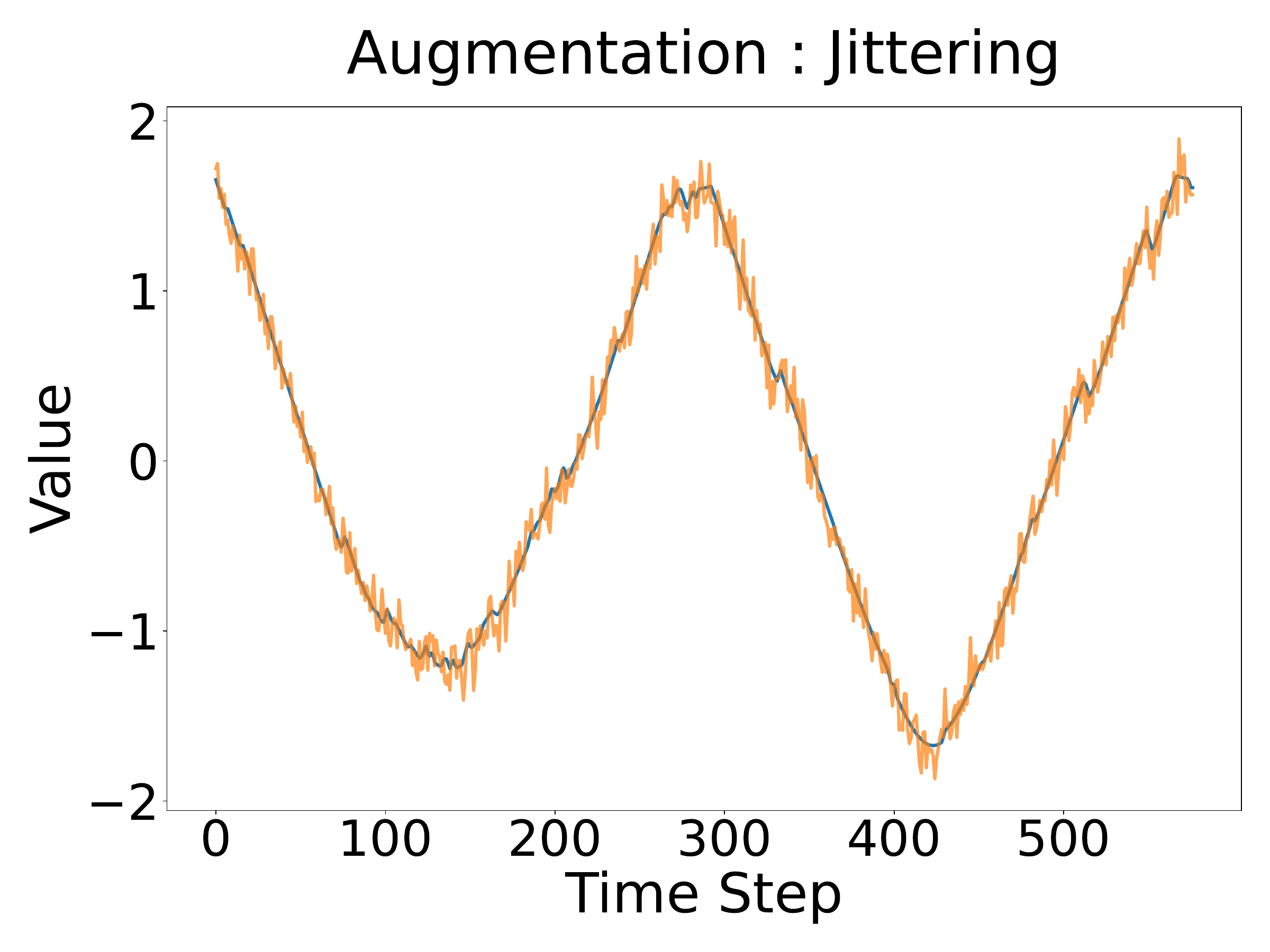}\hspace{3mm}
\includegraphics[width=0.3\textwidth]{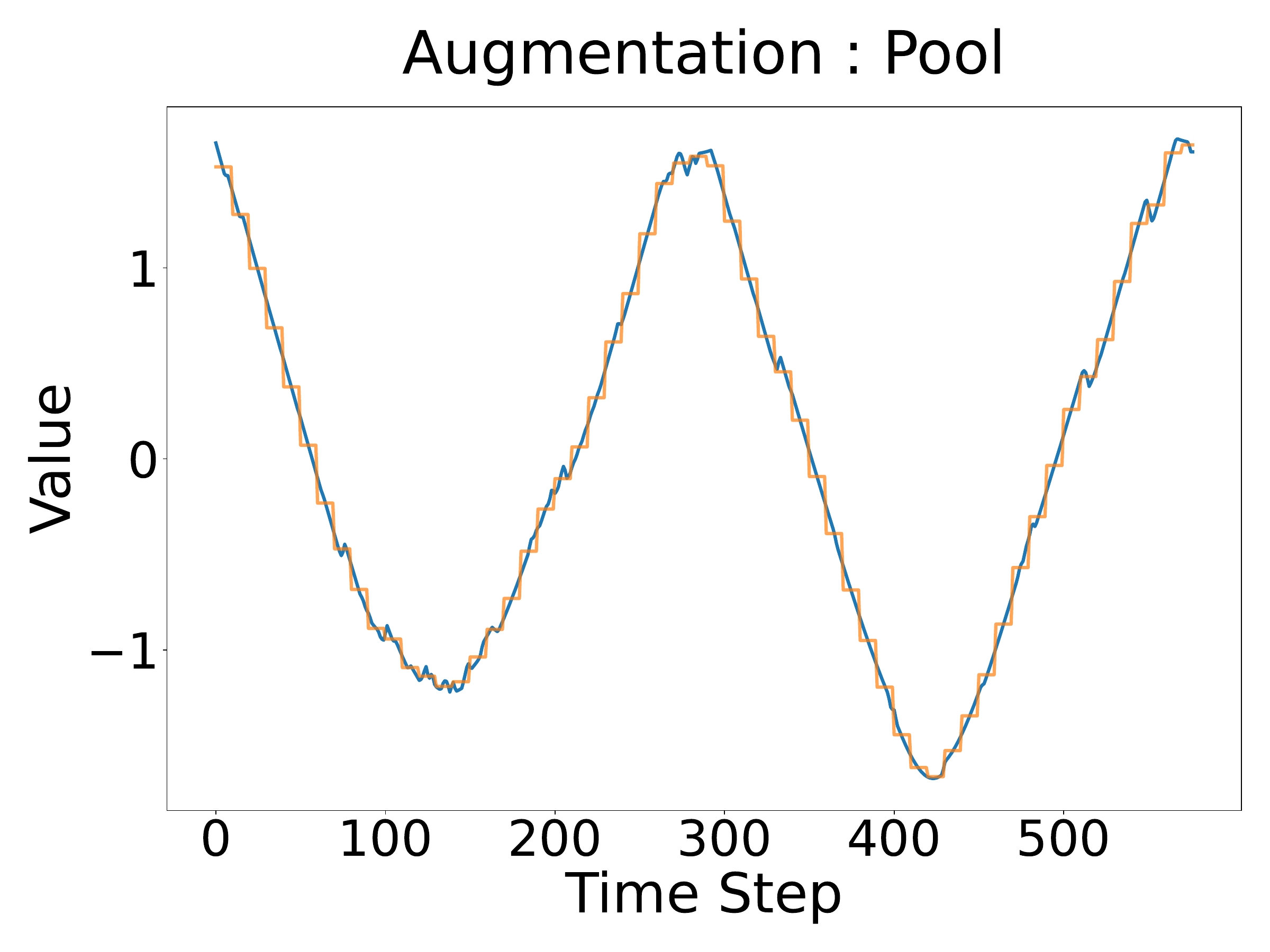}\hspace{3mm}
\includegraphics[width=0.3\textwidth]{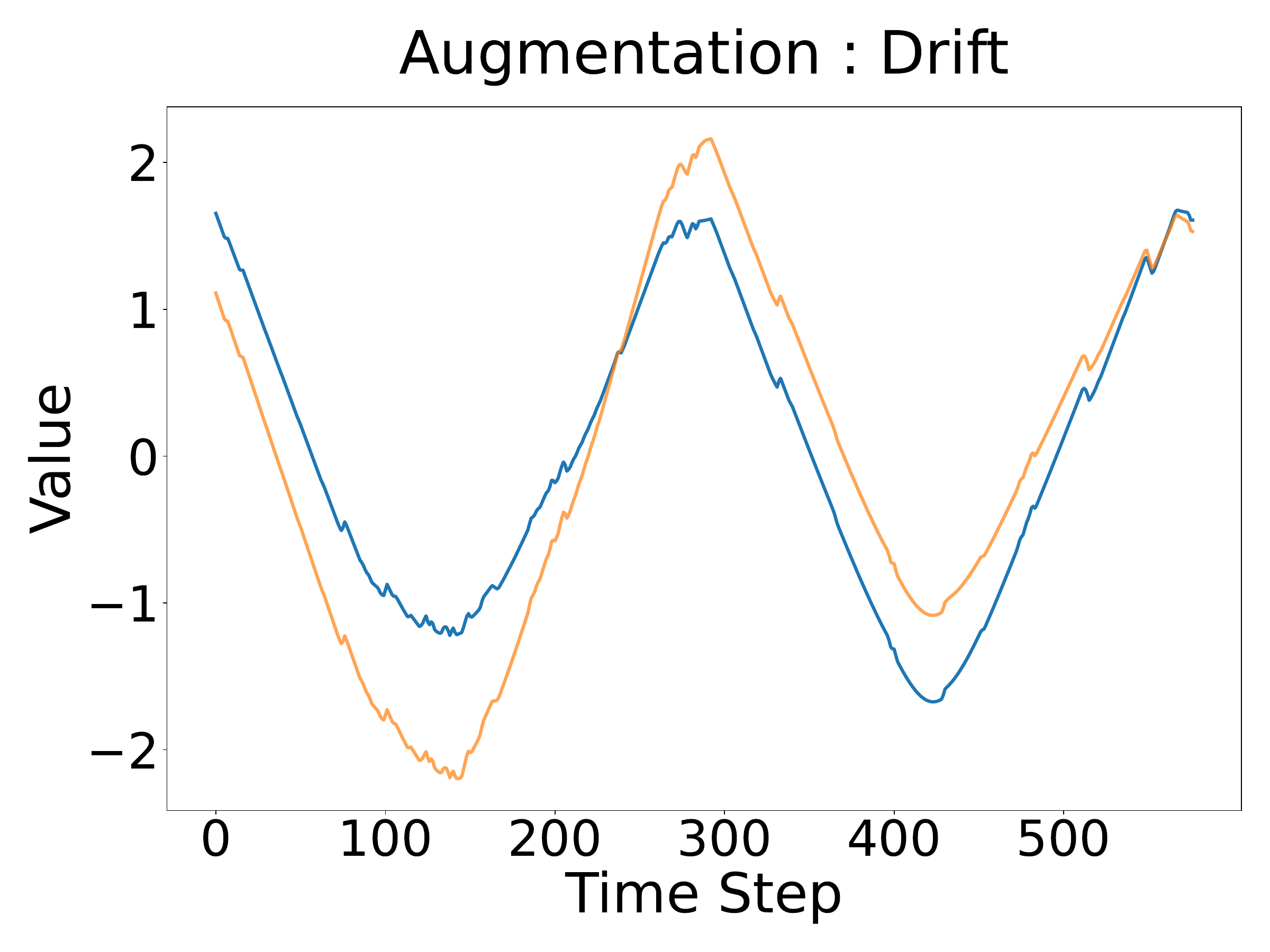}\\[4mm]
\includegraphics[width=0.3\textwidth]{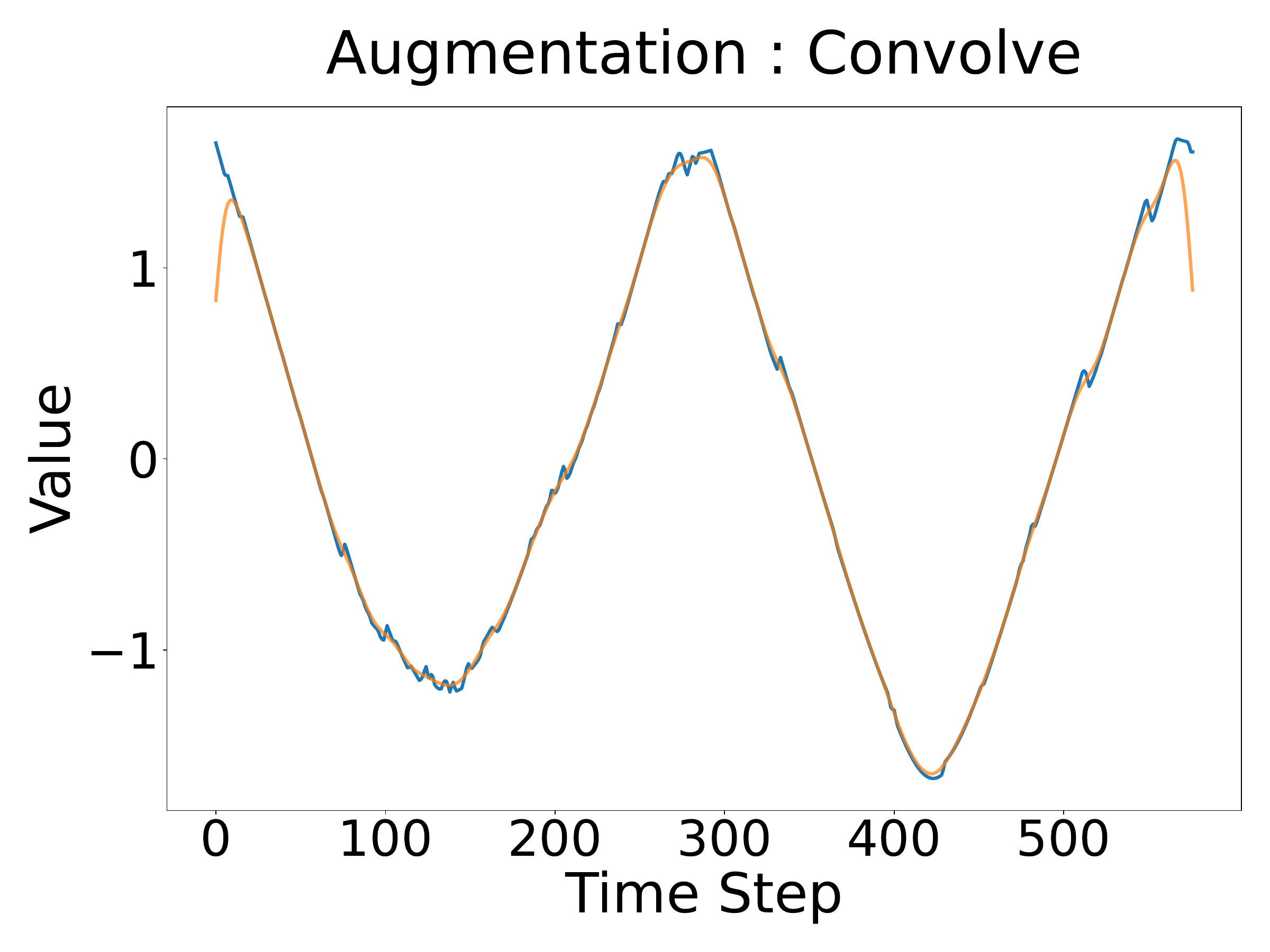}\hspace{3mm}
\includegraphics[width=0.3\textwidth]{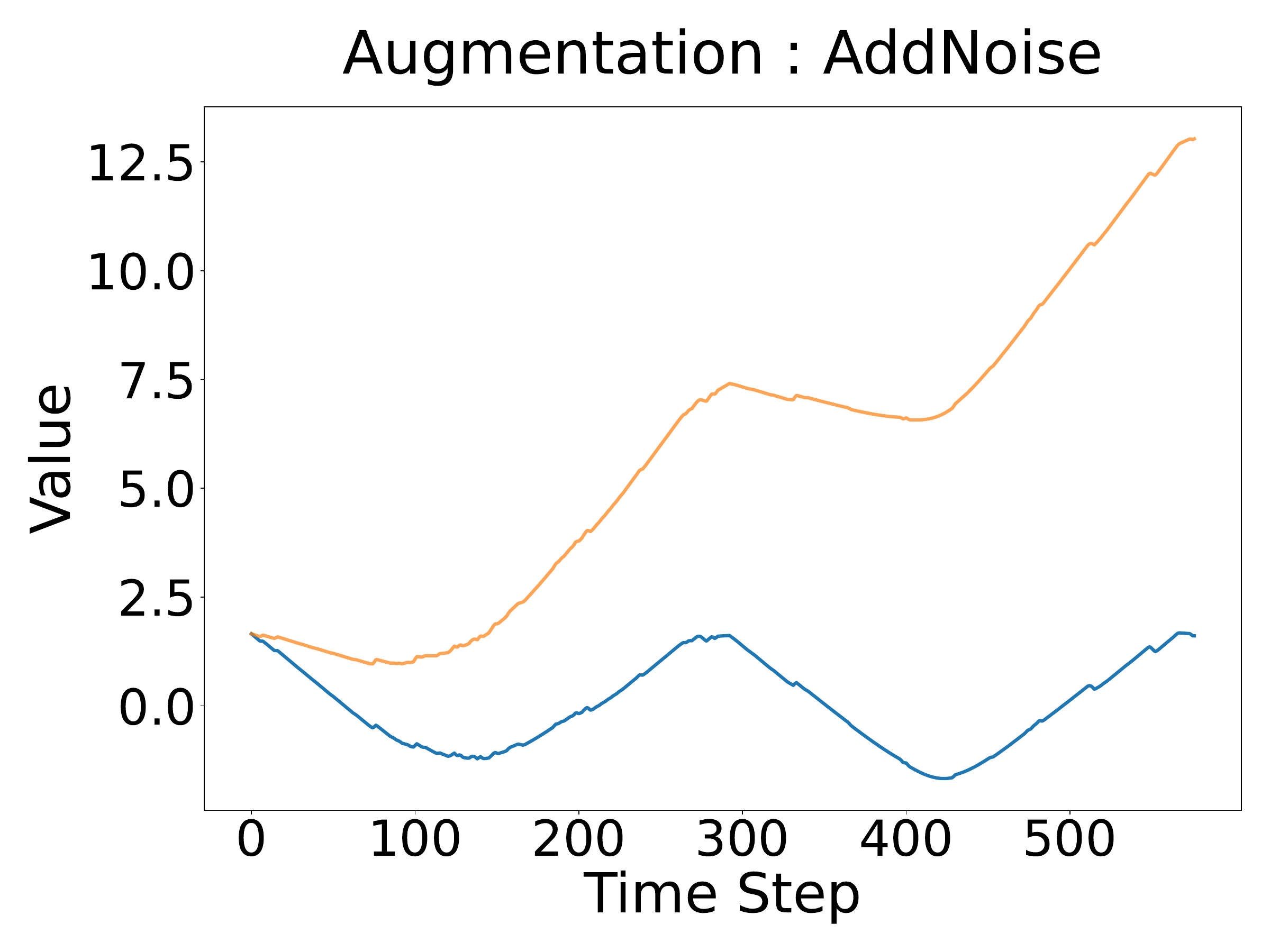}\hspace{3mm}
\includegraphics[width=0.3\textwidth]{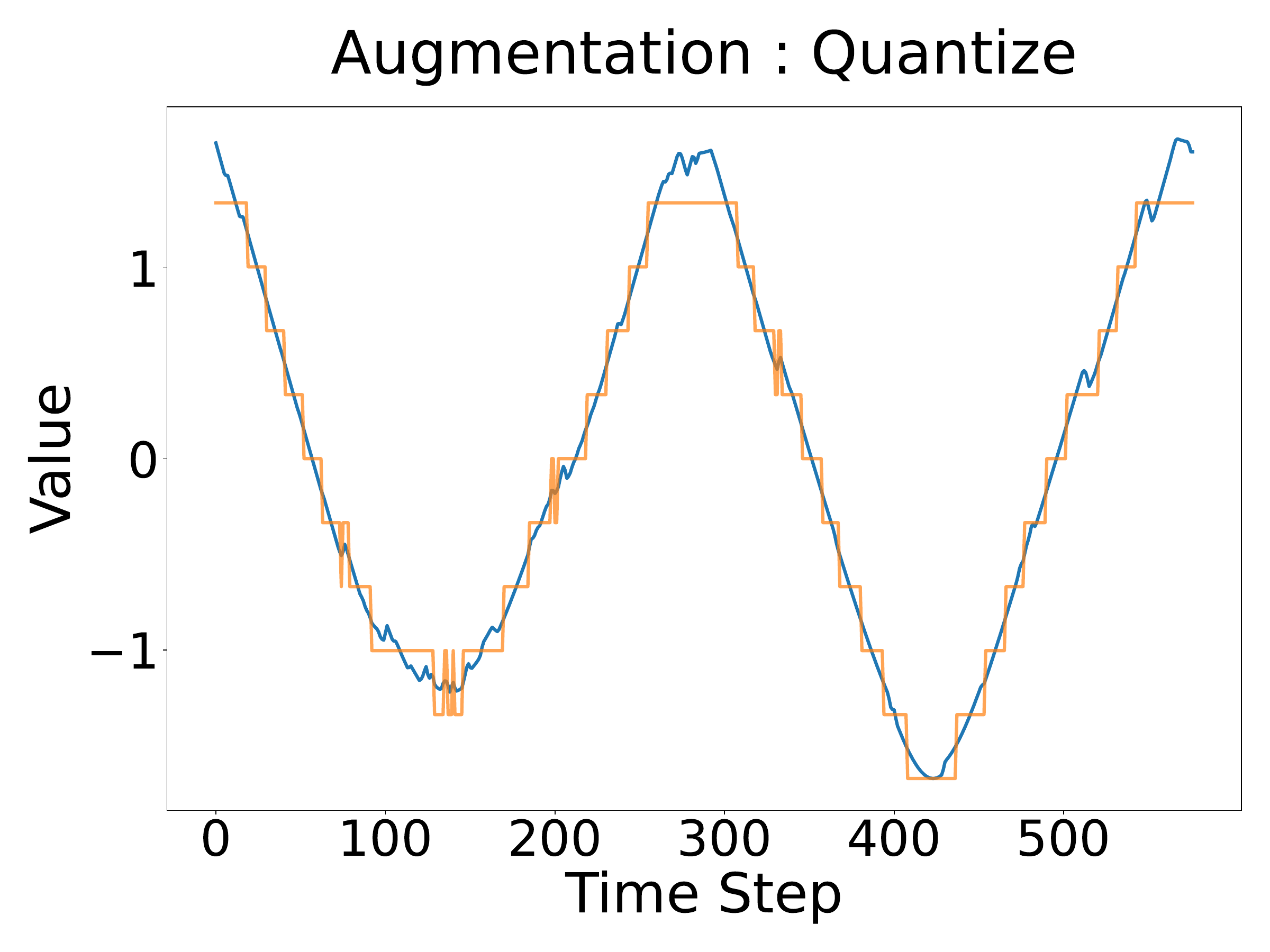}
\caption{Visualisation of some of the basic augmenters implemented in \gls{rats} on the Car dataset (first sample)
% \hh{shouldn't the top middle plot be labelled `Permute' (as this is the proper verb) or `Permutation' (the proper noun)? I am also puzzled by the fact that some of the augmentations are named using verbs and others using nouns ...}\ws{As I saw, Permutate is used more in the context of data augmentation as verbal form (\url{https://www.collinsdictionary.com/de/worterbuch/englisch/permutate}). The names here are matching the name of the functions in the codebase. I will take a note to fix the function names in the code base and then update the figure. We will either move to full verb or full nouns.}
}
\label{fig:basic_augmentations_grid}
\end{figure*}

For basic augmentations, we considered the most used augmentation methods in the literature.
\textcite{um2017data} presented \textbf{Rotation}, \textbf{Permutation}, \textbf{Scaling}, \textbf{Jittering}, and \textbf{Cropping} (also known as window Slicing) as valid transformations to time series data under which the labels are likely to remain invariant.
We also implemented an augmenter to inject noise into time series data.
For the types of noise, we followed \textcite{wen2020time} and considered uniform and Gaussian noise as well as injecting spikes or a slope trend into the data.
The other basic augmenters, such as \textbf{Reversing}, \textbf{Resizing}, or \textbf{Quantizing} time series, were inspired by \emph{tsaug}.
Figure~\ref{fig:basic_augmentations_grid} shows an overview of a sample time series from the Car dataset~\cite{dau2019ucrtimeseriesarchive} before and after basic transformations are applied.

\subsection{Frequency-Domain Transformations}\label{sec:freq_transforms}
While basic augmentations tweak samples directly in the time domain, frequency-domain transformations look at a signal ``as a whole'' by breaking it down into simpler periodic components.
In \emph{RATS}, two classic transformations are implemented, namely: \gls{dct} and \gls{fft}.

\paragraph{\acrfull{dct}}
works by rewriting a real-valued series as a weighted sum of cosine waves whose frequencies increase with the index of the respective component.  
 
\paragraph{\acrfull{fft}}
is an algorithm for computing the \gls{dft} that expresses a discrete-time signal as a sum of complex exponentials and enables its analysis in the frequency domain.
\gls{fft} powers almost every modern spectral analysis tool~\cite{cooley1965algorithm}.
% \hh{I find this description to be unsatisfactory. Surely there is a better and more accurate, yet still brief, explanation of FFT ...?}
% \ws{Done. I rewrote the definition}

% converts a signal into a list of complex sinusoids and powers almost every modern spectral analysis tool~\cite{cooley1965algorithm}. 

\medskip

After every forward–inverse transformation, the maximum absolute difference between the original and reconstructed datasets is checked.  
With a tolerance of \(10^{-8}\), both \gls{dct} and \gls{fft} reproduce every sample within a double-precision round-off.

% \ws{from here}
% so that they \hh{precisely who or what?} can be inserted anywhere in a pipeline without loss of information.
% \hh{that last sentence is very unclear; please try to reword for clarity.}
% \ws{I will remove this body of text. It is not really needed}

\subsection{Frequency Domain Augmentations}\label{sec:freq_augments}

In addition to the passive frequency-domain transformations presented in Section~\ref{sec:freq_transforms}, \emph{\gls{rats}} offers two active spectrum-level augmenters that inject variation directly in the frequency domain, namely: \gls{app} and Frequency Mask.
Both \gls{app} (fine-grained noise) and Frequency Mask (coarse band removal) offer complementary ways for nudging models toward spectral robustness without altering the overall length or label of the series.

\paragraph{\acrfull{app}}
adds zero-mean Gaussian noise to the magnitude and phase of every bin.  
\gls{app} includes the following steps:
\begin{enumerate*}[label=(\roman*), itemjoin={{, }}, itemjoin*={{, and }}]
    \item The series is first pushed through an \gls{fft} (if the input is still in the time domain);
    \item each complex coefficient is converted to polar form and perturbed by two independent normal distributions whose standard deviations the user can set;
    \item finally, it is mapped back to Cartesian coordinates, before an inverse \gls{fft} completes the process.
\end{enumerate*}

\paragraph{Frequency Mask}
suppresses a contiguous band of bins of a specific width, effectively simulating a narrowband dropout (notch filter).
The centre of the band is sampled uniformly at random, so different frequency regions are attenuated across a batch. 
The method works with both time‑domain inputs (via an internal spectral transform) and precomputed spectra, and it respects the global probability with which the augmenter is applied.

\subsection{Time Warping Augmentations}\label{sec:time_warping}
\begin{figure*}[t]
\centering
\includegraphics[width=0.4\textwidth]{plots/legend_strip.pdf}\\[2mm]
% Figure 2 (a)
\begin{minipage}{0.48\textwidth}
    \centering
    \includegraphics[width=\textwidth]{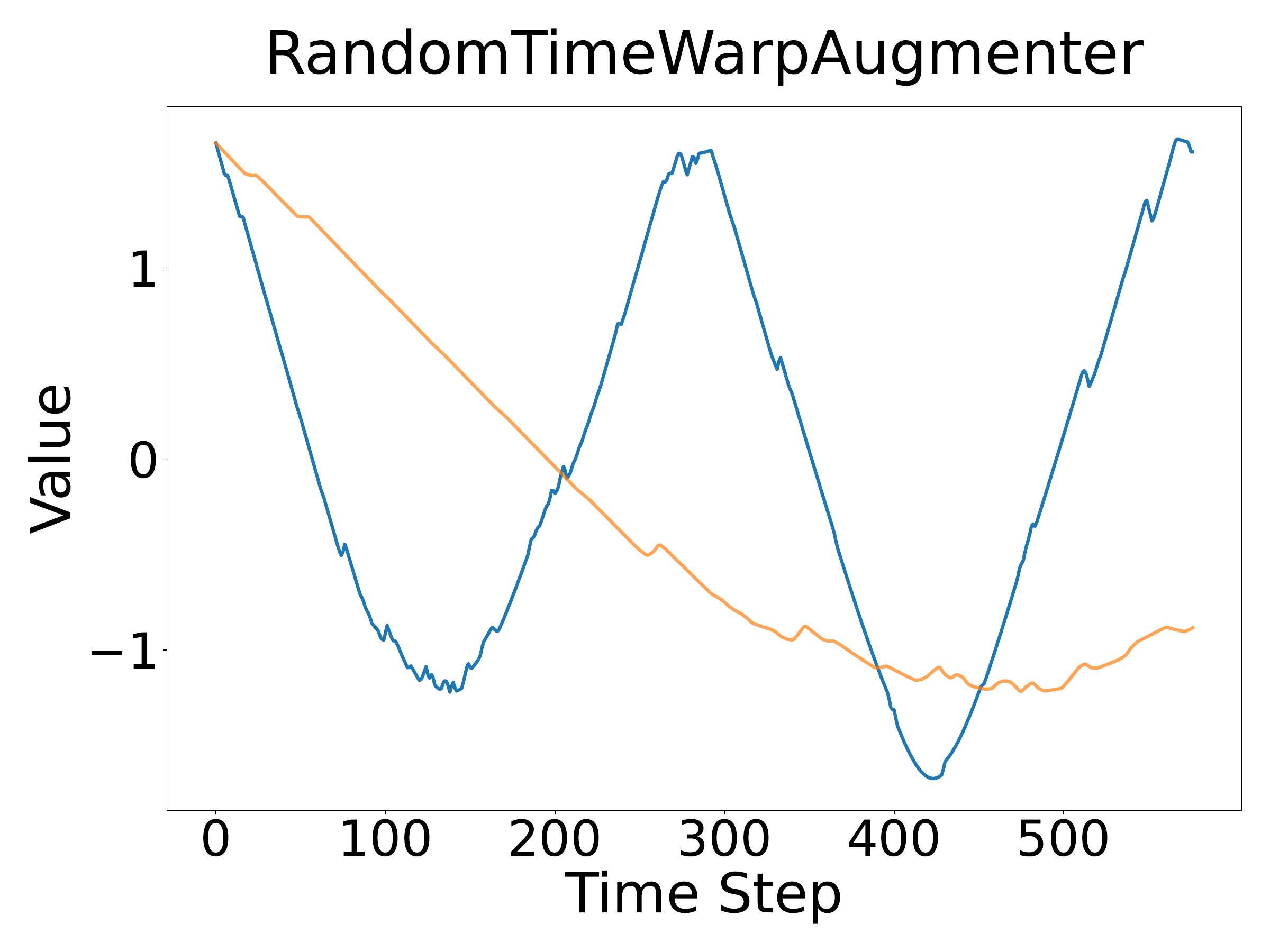}
    \subcaption{Random Time Warp  - Entire Series}
    \label{fig:rtw_window0}
\end{minipage}\hfill
% Figure 2 (b)
\begin{minipage}{0.48\textwidth}
    \centering
    \includegraphics[width=\textwidth]{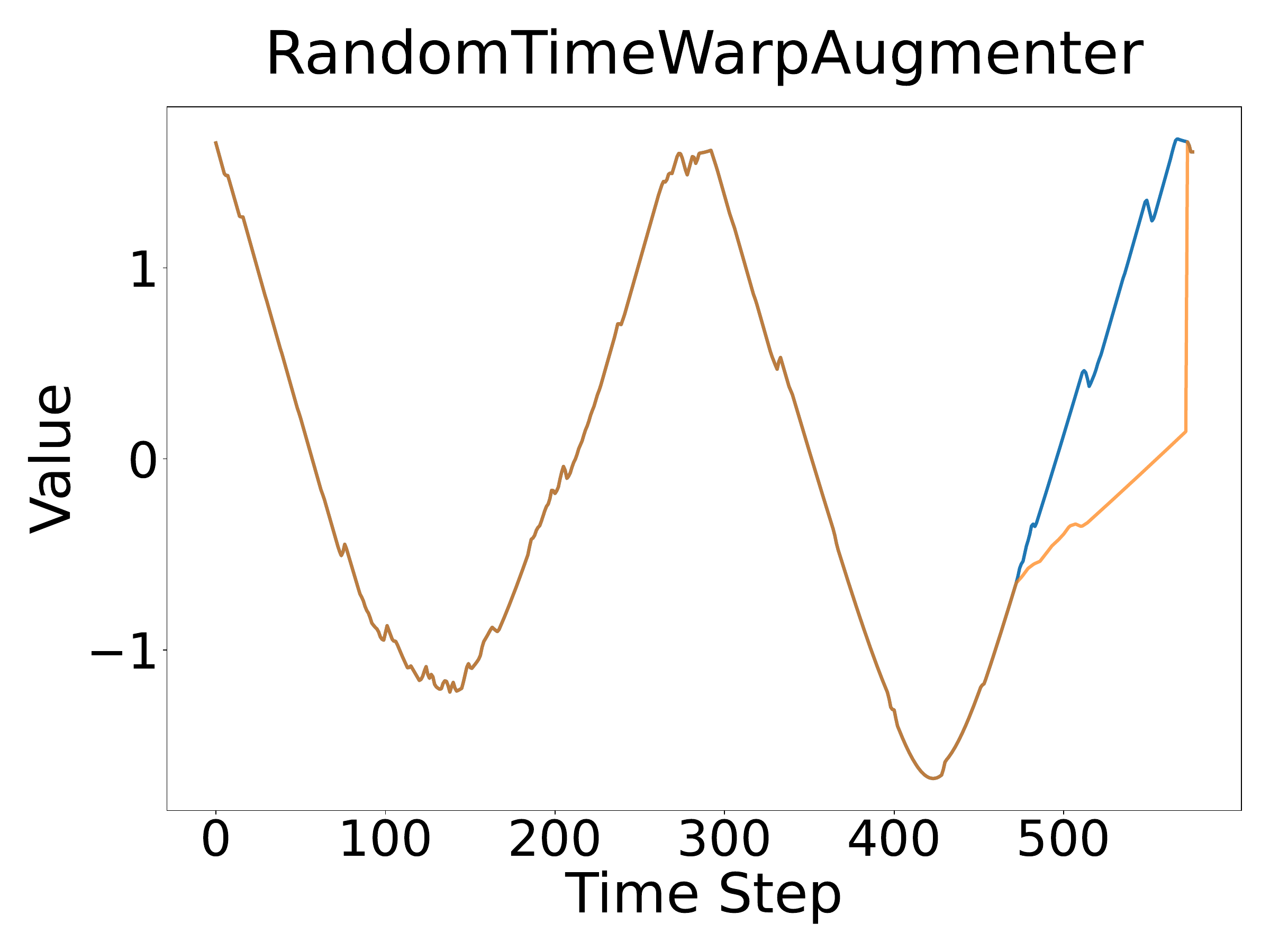}
    \subcaption{Random Time Warp - Size: 100}
    \label{fig:rtw_window100}
\end{minipage}

\caption{Time Warping Augmentation}
\label{fig:time_warping_comparison}
\end{figure*}

Time warping is a technique used to distort time in a time series.
The idea is to slow down or speed up the notion of time across the time series and distort its original representation.
This technique is commonly used to achieve dilation or contraction of audio signals~\cite{777905} and is therefore a very useful augmentation technique.
\emph{\gls{rats}} provides two variants of time warping:
\begin{enumerate*}[label=(\roman*), itemjoin={{, }}, itemjoin*={{, and }}]
\item for the entire time series sample (to represent noisy or incomplete signals)
\item for a specific, randomly selected window from the time series 
\end{enumerate*}.
Figure~\ref{fig:rtw_window0} shows a warped time series sample from the Car dataset \parencite{dau2019ucrtimeseriesarchive},
and Figure~\ref{fig:rtw_window100} shows a warped window of size 100 randomly selected from the same series sample.

\subsection{\acrshort{dtw} as a Similarity Measure}\label{sec:dtw}

To enable quality assessment of augmentations (further discussed in Section~\ref{sec:quality_assessment}) and to allow comparison of any two time series samples, we provide an implementation of the \gls{dtw} algorithm \parencite{itakura1975minimum, saoke78} as a similarity measure.
\gls{dtw} computes the distance between two time series samples by representing one samples as a warped form of the other, which gives a one-to-one mapping for every data point in the first sample to its corresponding data point in the other.

% Our implementation returns both, the distance normalized by the length of both the series (the length of the first series is used for the norm in case of uneven lengths) as well as the one-to-one mapping of the sample points in a separate vector. This especially helps in plotting and visualizing the dissimilarity or ``warp'' between the two series (Discussed in Section~\ref{sec:quality_assessment}).

\section{Implementation}\label{sec:implementation}

Rust offers memory safety and, together with the \texttt{rayon} crate, effortless data parallelism~\cite{bugden2022rustprogramminglanguagesafety}.  
Building on this foundation, \emph{\gls{rats}} processes entire datasets in milliseconds while still exposing a tidy, Python-friendly API (through \emph{RATSpy}).  
In this section, we outline how we utilised Rust features in practice to implement \emph{\gls{rats}} and \emph{RATSpy}.

% Because most everyday signals pack most of their energy into the first few DCT coefficients, this transform is a favourite for compression (e.g.,\ JPEG) and gentle denoising.  
% We use the high-performance \texttt{rustdct} planner: for each series we build a forward plan, copy the data into a scratch buffer, and run the transform in \(O(N\log N)\) time.  
% The inverse path calls the matching DCT-III and multiplies every value by\(2/N\).

% Our version relies on \texttt{rustfft}.  
% We build one plan for the current series length, lift each real series to a complex buffer (imaginary part set to 0), run the forward FFT, and finally flatten the output to an interleaved \texttt{[re\(_0\), im\(_0\), re\(_1\), im\(_1\), …]} vector—where \(\text{re}_i\) is the real part and \(\text{im}_i\) the imaginary part of the \(i\)-th frequency bin, so the result can still be in a simple \texttt{Vec<f64>}.

% Both these ways of time-warping are implemented under the \texttt{RandomTimeWarpAugmenter} and although the size of warped samples differs in both, their underlying implementation remains the same. Furthermore, to simplify the usage of this augmenter, we default to augmenting the entire time-series in case an invalid window size, such as 0 or a number greater than the size of the dataset is provided.

\paragraph{Parallelisation.}\label{sec:parallelisation}
Almost every augmenter in \emph{\gls{rats}} can run in parallel.  
Activating the \texttt{parallel} flag switches each internal loop from \texttt{Iterator::for\_each} to \newline \texttt{rayon::ParallelIterator::for\_each}.  
This change enables \texttt{rayon} to schedule one logical task per series and execute these tasks concurrently on a global, work-stealing thread pool. 
When a pipeline receives a batch with \texttt{parallel=true}, every component that supports parallelism processes its slice concurrently; components that do not support it fall back to their serial path without affecting the rest of the chain. Therefore, users need to set the flag only once at the top level.
The only exception to parallelism is the \texttt{Repeat} augmenter.  
Repeating a series \(n\) times requires cloning memory, which violates the borrowing rules of Rust when multiple mutable references exist.  
Consequently, \texttt{Repeat} always executes serially.
Because the scheduler of \texttt{rayon} relies on work-stealing, idle threads dynamically \emph{steal} unfinished series from busy threads, maintaining high utilisation with minimal contention. 

\paragraph{Pipelining.}\label{sec:pipelining}
All augmenters implement a common trait that contains the \texttt{augment\_one} and \texttt{augment\_batch} methods.
A pipeline is an ordered list of such objects plus a dispatcher that runs through them.  

% \hh{the following is imprecise; I guess what you mean is that with each augmenter in the list, a probability is associated, and when executing the pipeline, the augmenter is executed with that probability, and otherwise, the sample is passed unchanged?}\ws{Modified}

For every batch element, the dispatcher checks the probability of each augmenter in the list when executing the pipeline.
The augmenter is executed with it associated probability; otherwise, the sample is passed unchanged.  
Two pipeline execution options are available:
% carries out a random draw;
% if the draw is positive, the augmenter is executed in place

\begin{enumerate*}[label=(\roman*), itemjoin={{, }}, itemjoin*={{, and }}]
    \item {\emph{Standard mode}}: the entire batch is augmented sequentially  
    \item {\emph{Per-sample mode}}: enabled via the \texttt{per\_sample} flag.  
          Here, we first split the batch into slices and then run the entire chain on each slice in its own worker thread, so every time series sample is augmented independently.
\end{enumerate*}
Because every augmenter conforms to the same interface, users can mix basic, frequency-domain, and warping augmenters in any order.

\paragraph{Python Bindings.}\label{sec:bindings}

The bindings were written using the Python-Rust binding library \texttt{PyO3}~\cite{pyo3}. 
They work by using wrapper objects that contain objects from \emph{\gls{rats}} and providing an interface to access them indirectly.
In most cases, this works; however, it cannot bind the \texttt{AugmentationPipeline} augmenter, because \texttt{PyO3} does not support the binding of dynamically dispatched types using the \texttt{dyn} keyword;
therefore, this augmenter was re-implemented in \emph{RATSpy} using Python.
This does not have a significant impact on the performance, because the pipeline calls only other augmenters that are implemented in Rust.
Implementing the per-sample pipelining execution method for Python bindings was not necessary, because of the negligible performance differences between the two pipelining methods, i.e., per-dataset and per-sample (see Section~\ref{sec:time_benchmarks}).

\begin{figure*}[t]
\centering
\includegraphics[width=0.4\textwidth]{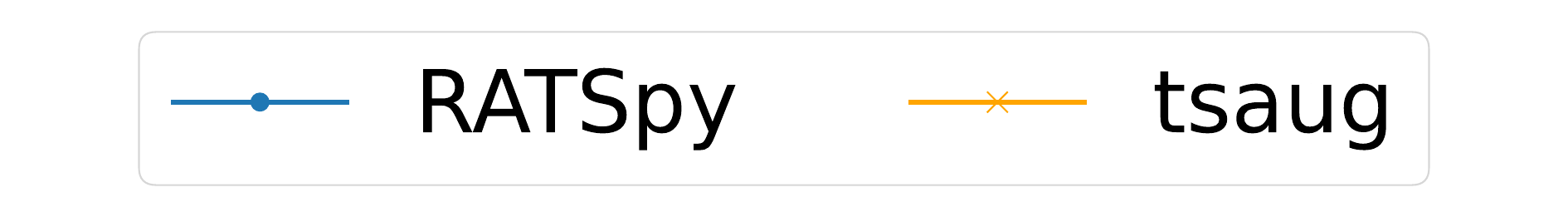}\\[2mm]
\begin{minipage}{0.48\textwidth}
    \centering
    \includegraphics[width=\textwidth]{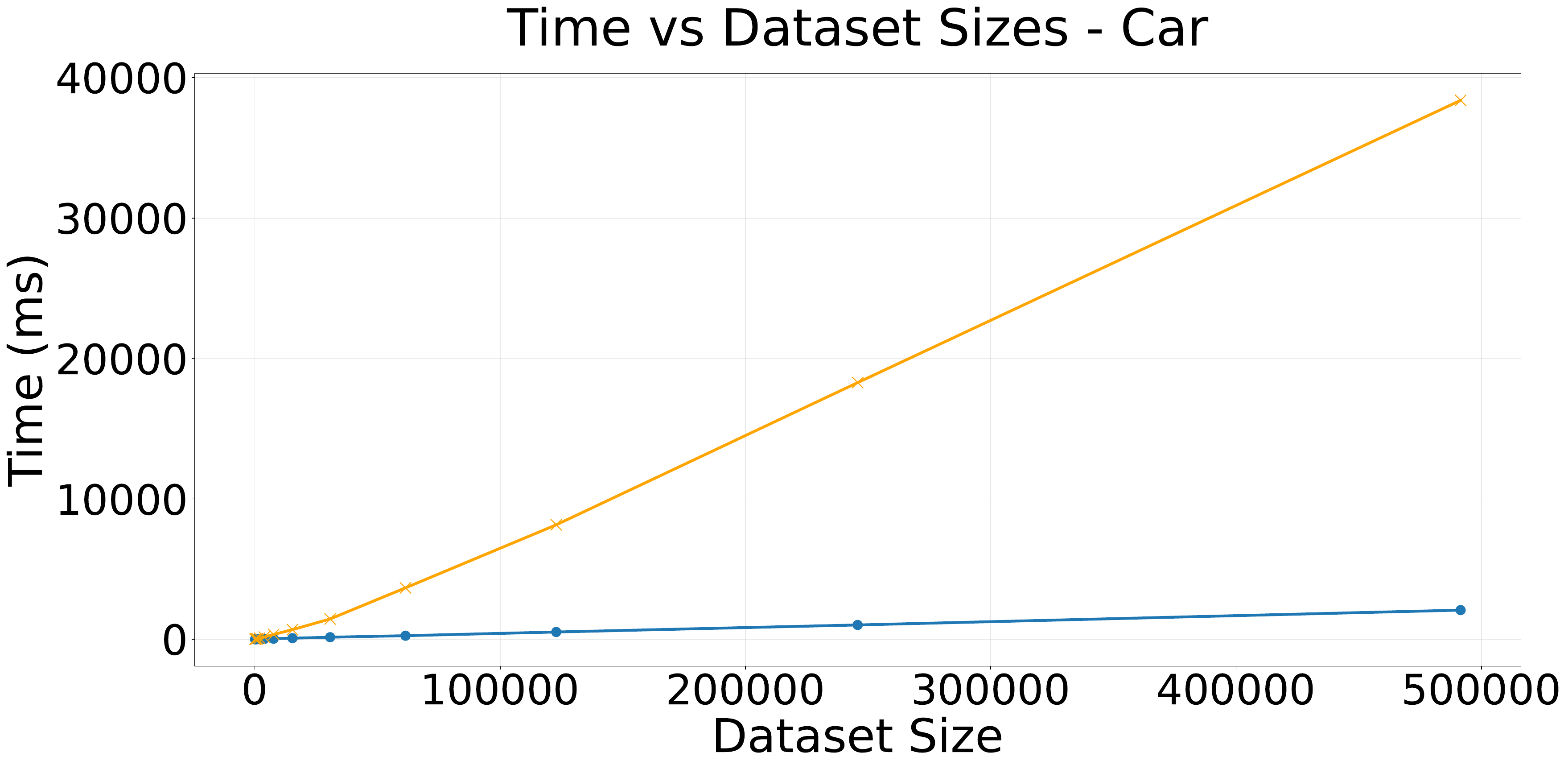}
    \subcaption{RATSpy vs tsaug - Time Benchmarking for up to 500k samples}
    \label{fig:car_time_size}
\end{minipage}\hfill
\begin{minipage}{0.48\textwidth}
    \centering
    \includegraphics[width=\textwidth]{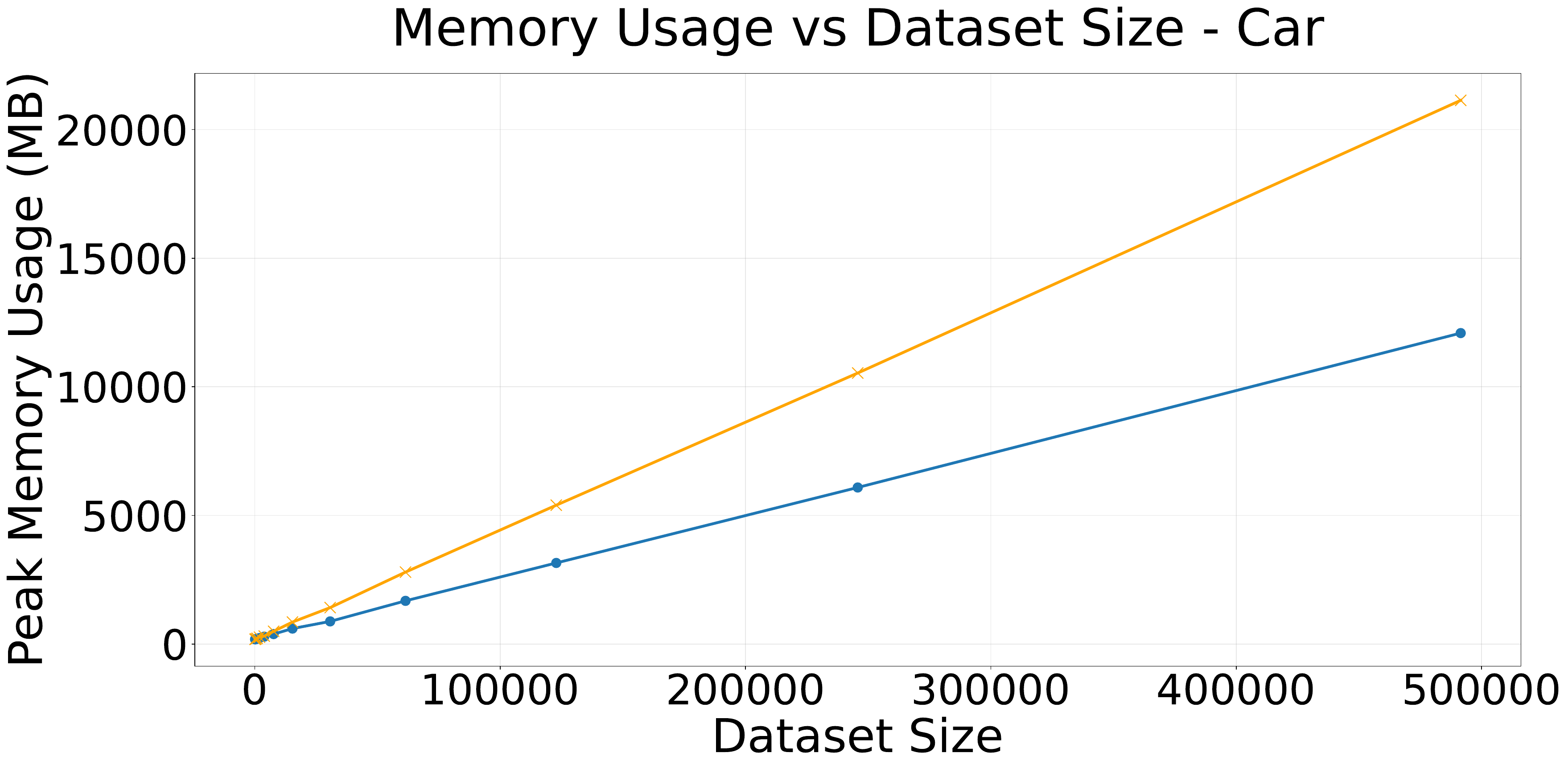}
    \subcaption{RATSpy vs tsaug - Memory Benchmarking for up to 500k samples}
    \label{fig:car_mem_size}
\end{minipage}

\caption{Plots of runing time (ms) and memory (MB) \emph{vs} dataset size for RATSpy and tsaug}
\label{fig:time_memory_usage_comparison}
\end{figure*}

% Scatter plot
%\begin{figure*}[t]
%\includegraphics[width=0.48\textwidth]{plots/benchmarking/Car_time_vs_memory_scatter.eps}
%\centering
%\caption{\centering  Scatter Plot of Time (s) vs Memory (MB) for RATSpy and tsaug augmenters \todo{make text bigger in the figure}}
%\label{fig:car_time_mem}
%\end{figure*}

\begin{figure*}[t]
\centering
\includegraphics[width=0.4\textwidth]{plots/legend_strip.pdf}\\[2mm]
\includegraphics[width=0.93\textwidth]{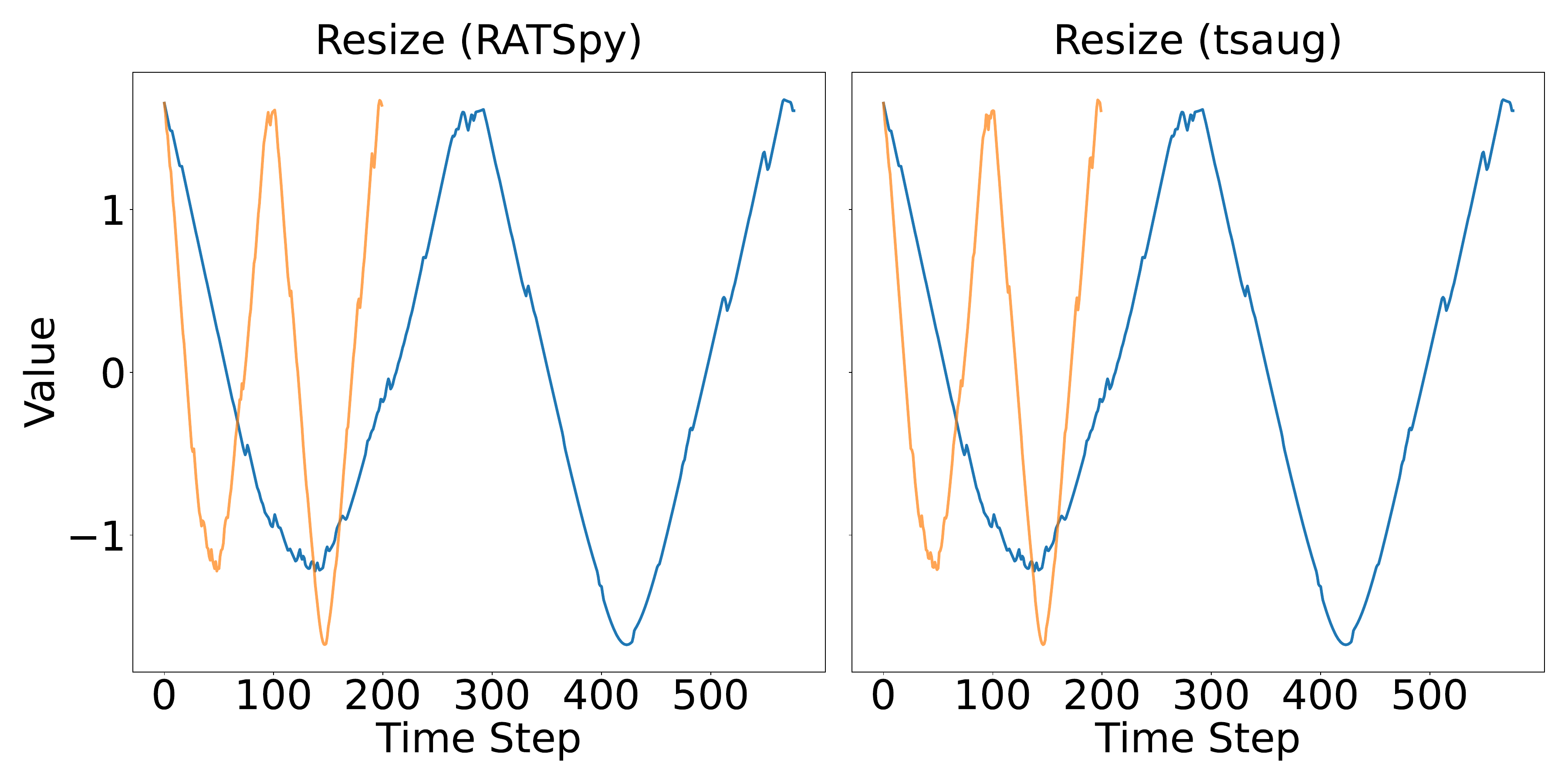}
\caption{\centering Comparison of time series augmented using RATSpy and tsaug}
\label{fig:resize_tsaug_pyfraug}
\end{figure*}

\begin{figure*}[t] 
\centering
\includegraphics[width=0.4\textwidth]{plots/legend_strip.pdf}\\[2mm]
\includegraphics[width=0.48\textwidth]{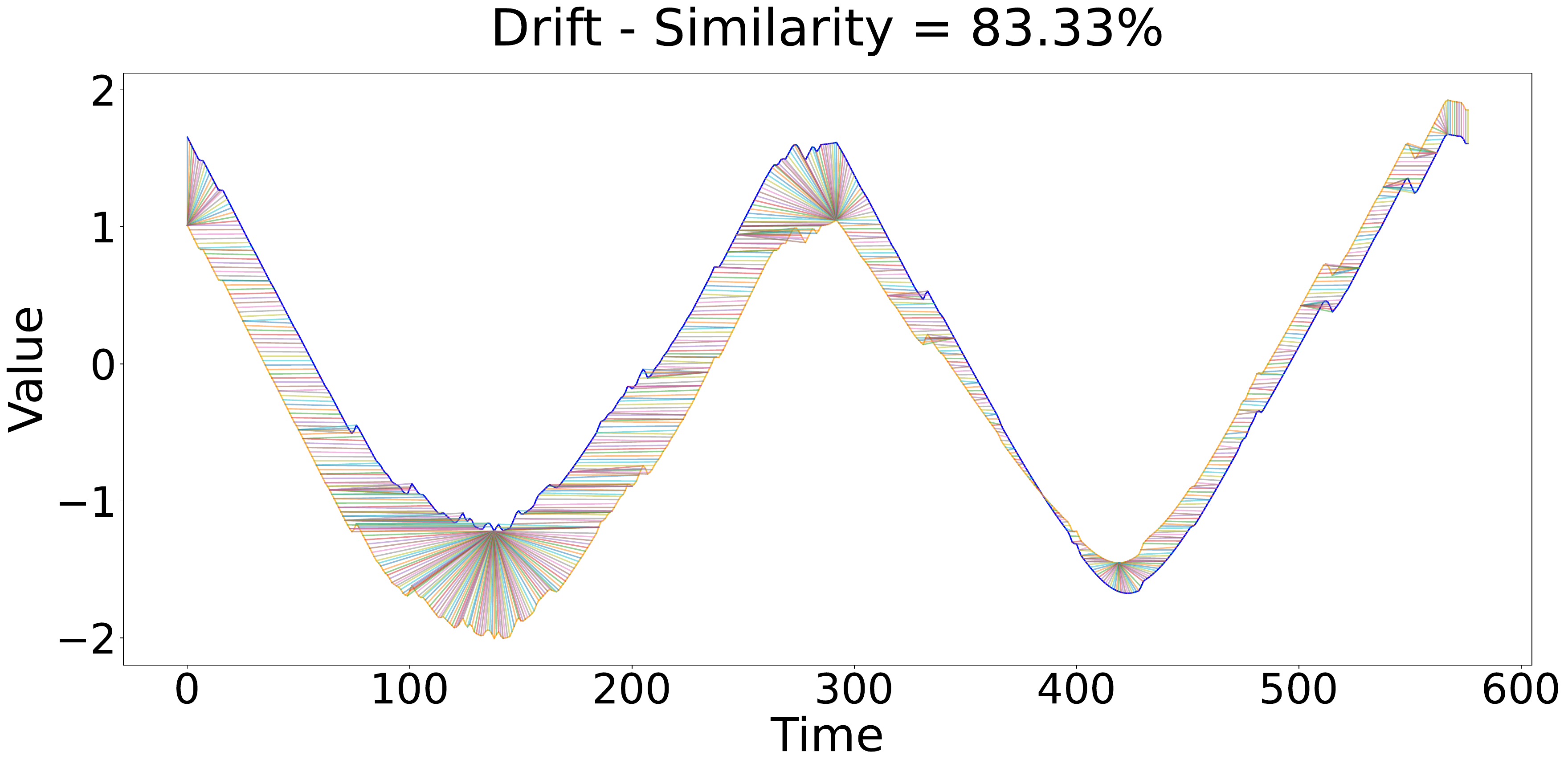}\hspace{1mm}
\includegraphics[width=0.48\textwidth]{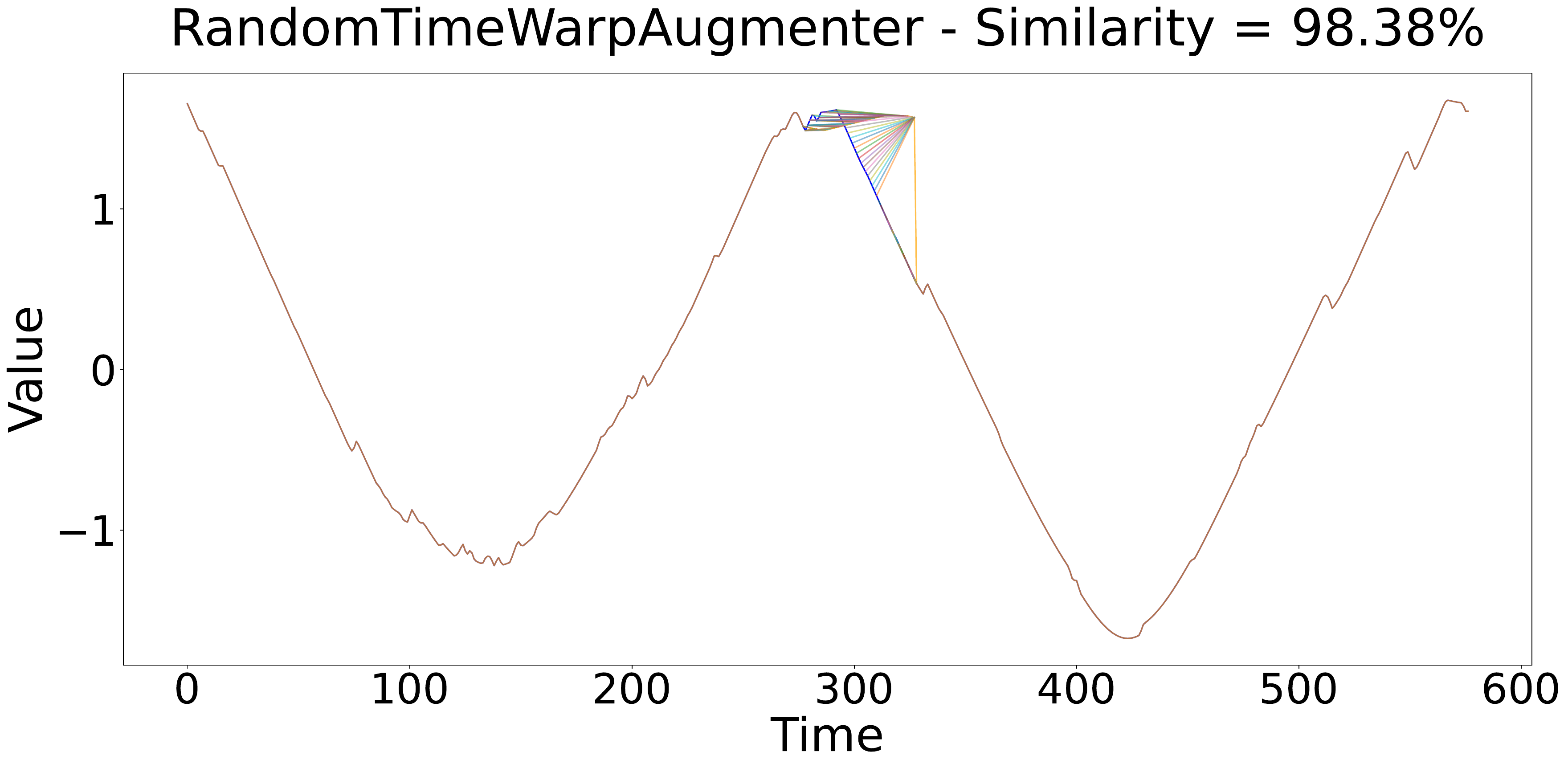}\\
[4mm]
\includegraphics[width=0.48\textwidth]{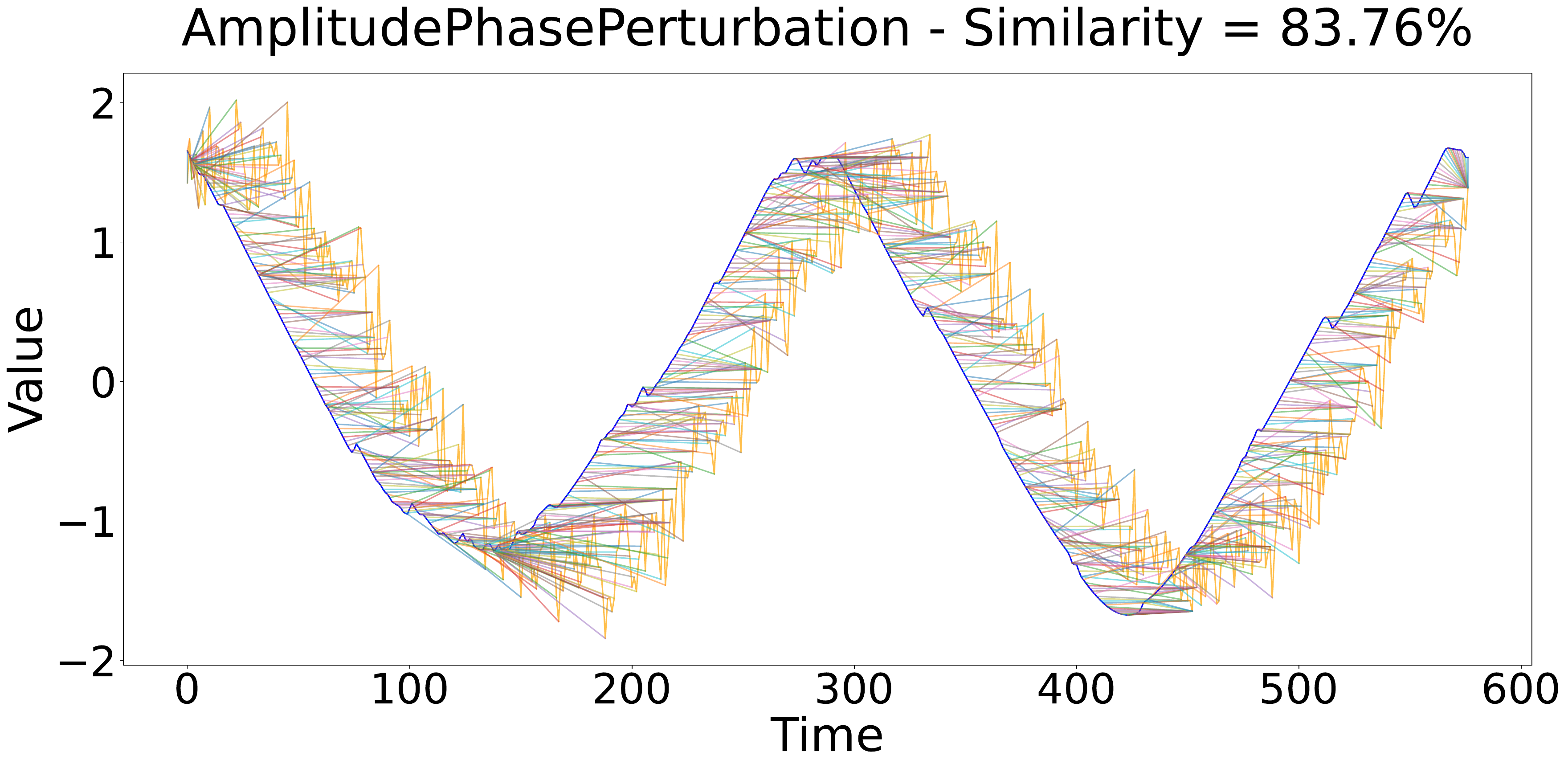}\hspace{1mm}
\includegraphics[width=0.48\textwidth]{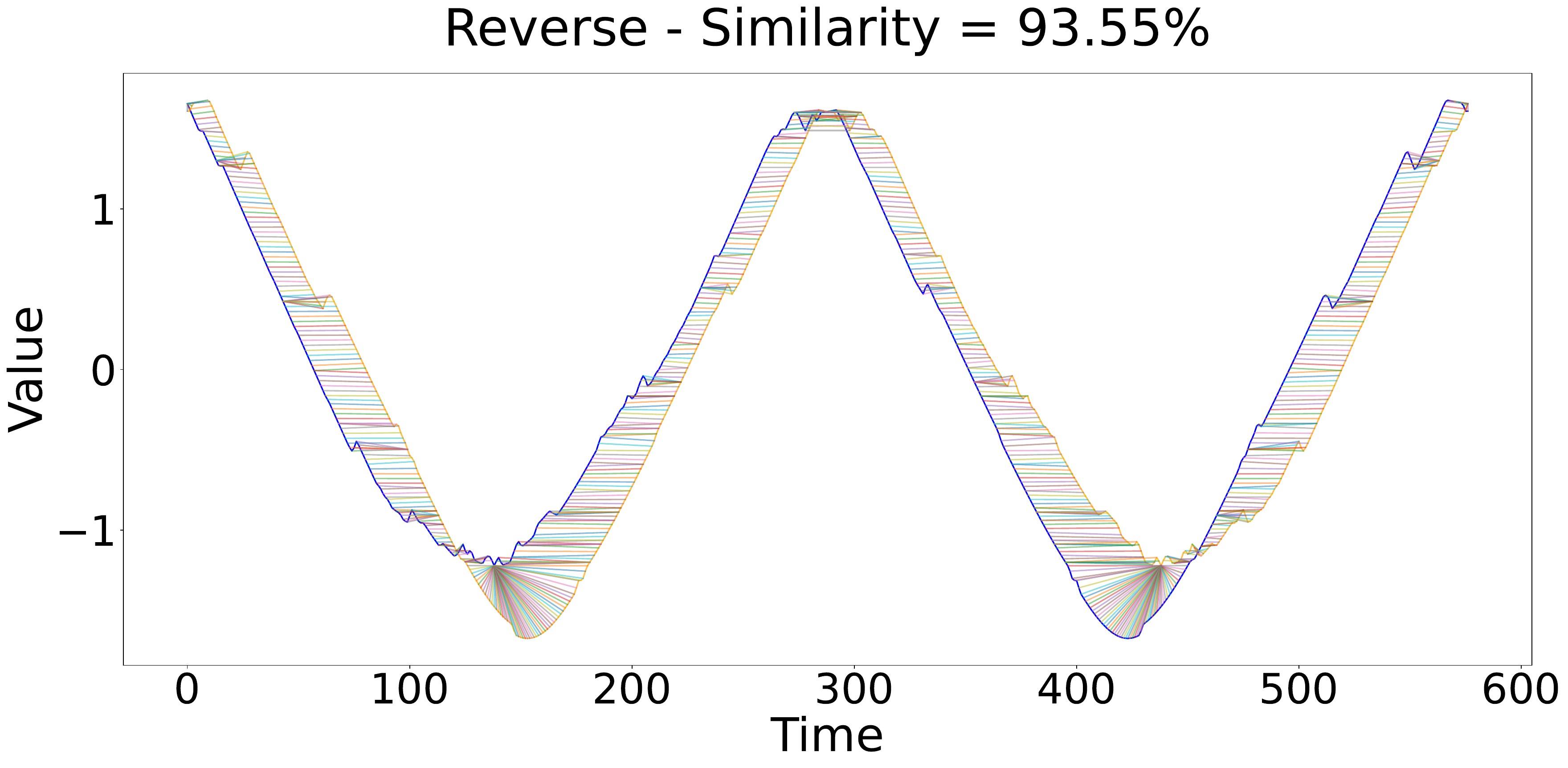}

\caption{\centering Dynamic Time Warping Analysis for Drift, Random Time Warp, \gls{app}, and Reverse augmentations (top left to bottom right)}
\label{fig:dtw_plots}
\end{figure*}

\section{Performance Evaluation}\label{sec:perf_eval}

To assess the performance of our Rust library 
% \hh{previously, you call it a library ... perhaps it would be good to make a connection between the two terms here or earlier}
\emph{\gls{rats}} and its Python counterpart \emph{RATSpy}, we used the popular Python-based time series augmentation library \emph{tsaug} \cite{tsaug} as a baseline.

\paragraph{Setup of Experiments.}
To benchmark the performance of \emph{RATSpy} against \emph{tsaug}, we used all univariate equal-length datasets from the aeon toolkit~\parencite{middlehurst_2024_aeon}, a total of 143 datasets (see Appendix \ref{appsec:detailed_results} for details).
We measured the per-augmenter running time for both libraries and recorded the peak memory usage for each augmenter, using the \emph{memory profiler} hook in Python.
We then constructed a pipeline that consists of the full augmentation chain common to \emph{RATSpy} and \emph{tsaug}, and evaluated it using the same datasets for both time and memory performance. 

\subsection{Time Benchmarking}\label{sec:time_benchmarks}
The time benchmarking results demonstrated that \emph{RATSpy} outperformed \emph{tsaug} by a significant margin across all five datasets, as shown in Table~\ref{tab:pipeline_benchmarks_all}.
For the largest dataset, \emph{Sleep}, 
% \hh{slightly modified -- check:}
running the full \emph{RATSpy} pipeline takes about $2.2$ seconds, while the same pipeline in \emph{tsaug} requires more than 21 seconds to execute.
Across all datasets, \emph{RATSpy} is at least 45.3\% (\emph{ECGFiveDays}) faster than \emph{tsaug} and at most 94.8\% (\emph{RightWhaleCalls}) faster.
On average, the implementation in \emph{\gls{rats}} is approximately 74.4\% faster than that in \emph{tsaug}.
Another key observation is that execution time of \emph{tsaug} for augmentation increases exponentially with dataset size, whereas for \emph{RATSpy}, the running time remains consistent with no major increases.
% \hh{what precisely does `almost' mean? can you add a sentence or footnote to clarify this?}
Figure~\ref{fig:car_time_size} shows the comparison of the execution time of the augmentation pipeline for both \emph{RATSpy} and \emph{tsaug} as a function of dataset size.

We also tested whether the chosen pipelining method has an impact on the performance.
For this, we executed a pipeline with seven different augmenters on the \emph{Sleep} dataset.
For the standard execution method, where augmenters are applied sequentially on the entire dataset, the pipeline finished the augmentation in $26.8$ seconds.
The other execution method is a ``per-sample'' methods, where the augmentations are applied on one sample at a time rather than on the entire dataset.
The per-sample execution method took $28.6$ seconds;
therefore, the performance difference is negligible.

% font size 16 in matplotlib
\begin{figure}[t]
\centering
    \includegraphics[width=0.35\textwidth]{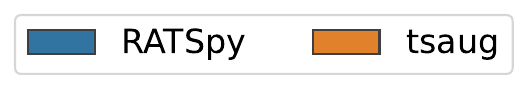}\\
    \begin{minipage}{0.95\textwidth}
        \centering
        \includegraphics[width=\linewidth]{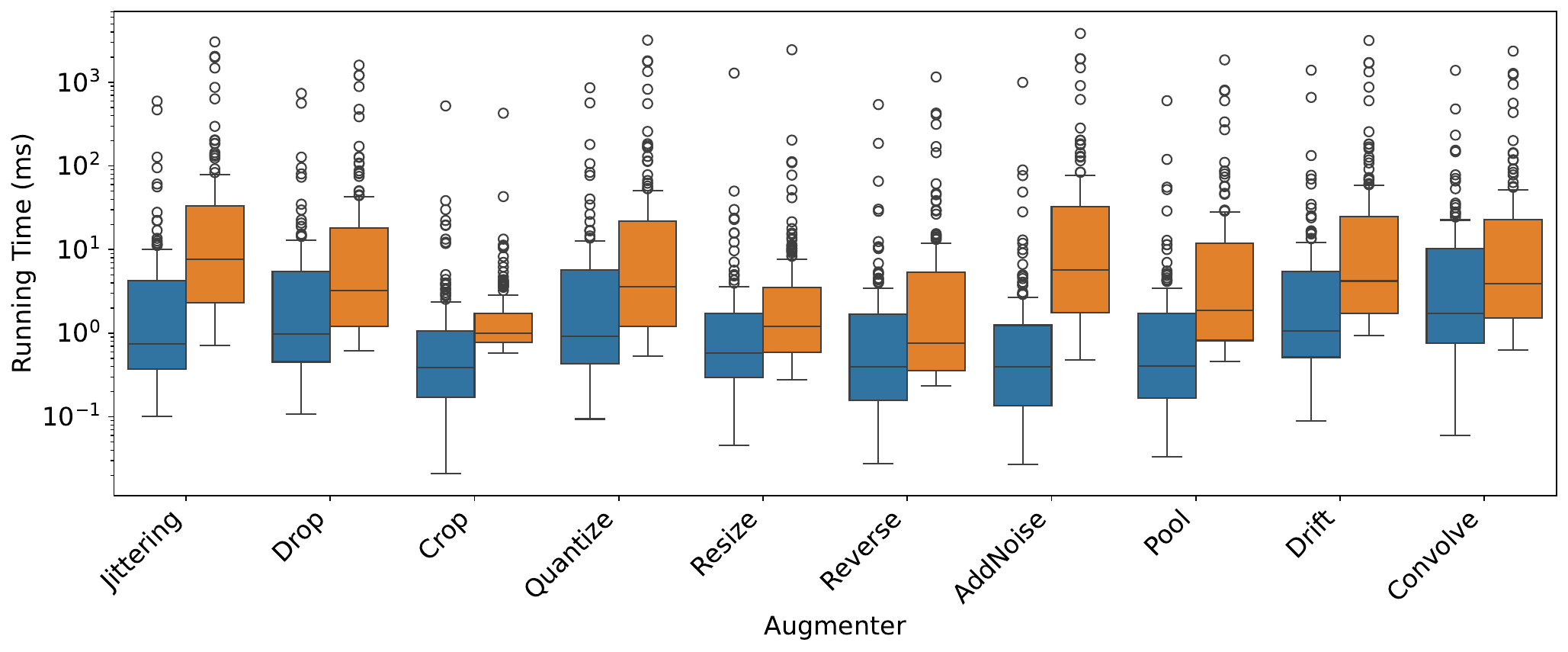}
        \subcaption{Per-Augmenter Running Time Benchmark (All Datasets)}
        \label{fig:time_box_plot}
    \end{minipage}
    
    \begin{minipage}{0.95\textwidth}
        \centering
        \includegraphics[width=\linewidth]{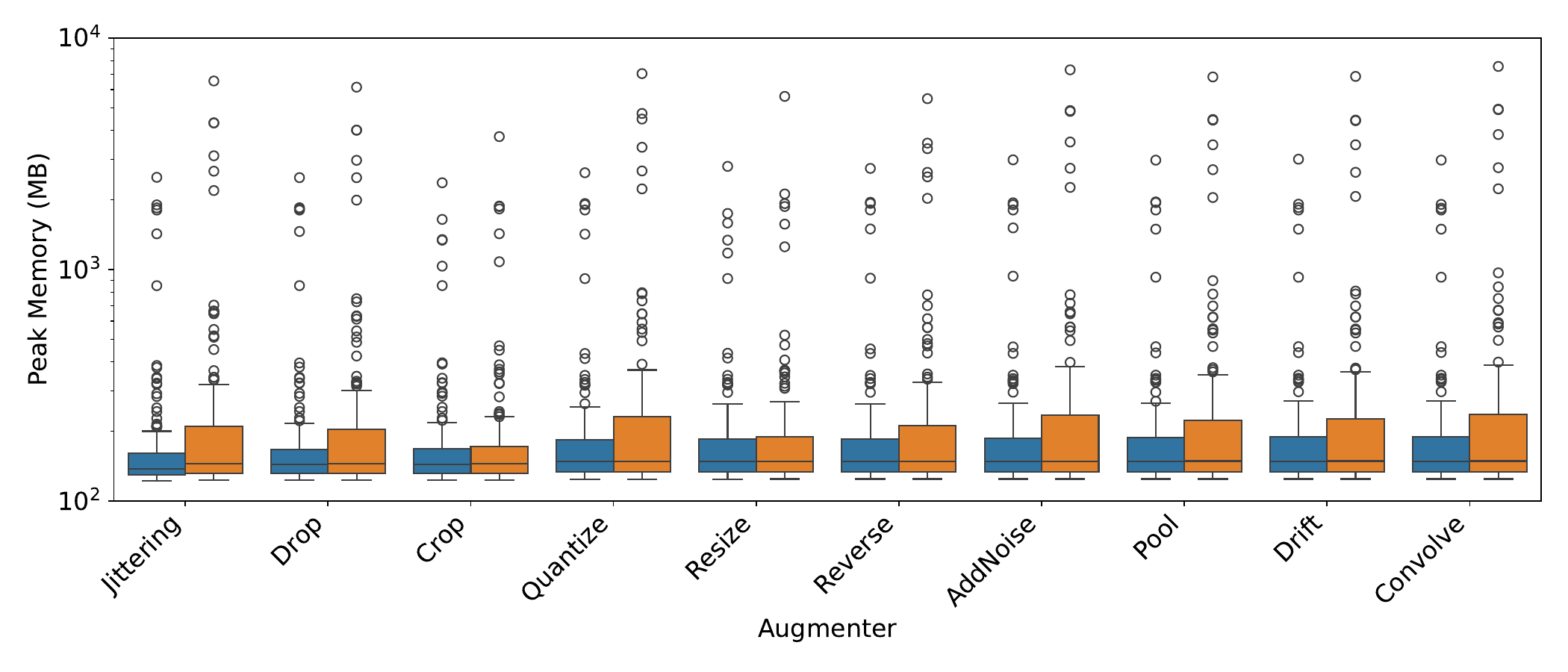}
        \subcaption{Per-Augmenter Peak Memory Usage Benchmark (All Dataset)}
        \label{fig:memory_box_plot}
    \end{minipage}
    
    \caption{Box plots of per-augmenter metrics for both running time and peak memory usage benchmarks}
    \label{fig:box_plots}
\end{figure}

The benchmarking results for an example augmentation pipeline are presented in Table~\ref{tab:pipeline_benchmarks_all} and illustrated for each augmenter in Figure \ref{fig:time_box_plot}; it can be observed that \emph{RATSpy} consistently outperforms \emph{tsaug}.

\subsection{Memory Benchmarking}\label{sec:mem_benchmarks}
The peak memory usage profiling shows a similar trend as that of the time-benchmarking (see Table~\ref{tab:pipeline_benchmarks_all}).
On the largest dataset (\emph{Sleep}), the example augmentation pipeline using \emph{RATSpy} peaked at
\(4.83\;\mathrm{GB}\), while 
\(9.26\;\mathrm{GB}\) were required when using \emph{tsaug}.
For small datasets (e.g., \emph{ArrowHead}), both \emph{RATSpy} and \emph{tsaug} required the same amount of memory.
For large datasets, \emph{RATSpy} outperformed \emph{tsaug}.
\emph{RATSpy} showed the largest improvement compared to \emph{tsaug} on the dataset \emph{Sleep}, requiring 47.9\% less peak memory.
On average, \emph{RATSPpy} used approximately 7.3\% less memory than \emph{tsaug}.
A key observation here is that, while on smaller datasets such as Car (120 samples), the memory usage of both libraries is similar (approximately \(100\;\mathrm{MB}\)), because the Python interpreter and NumPy already occupy the same baseline memory, substantial differences arise as dataset size increases.
A linear trend can be observed for peak memory usage as a function of dataset size; however, the slope for \emph{tsaug} is noticeably steeper (see, e.g., Figure \ref{fig:time_memory_usage_comparison}), which suggests extra temporary memory allocations in pure Python compared with the disciplined garbage collection mechanisms in Rust.

Benchmarking results for an example augmentation pipeline are presented in Table~\ref{tab:pipeline_benchmarks_all} and illustrated for the individual augmenters in Figure \ref{fig:memory_box_plot}.
It can be observed that \emph{RATSpy} is consistently more memory efficient than \emph{tsaug}.

\subsection{Quality Assessment}\label{sec:quality_assessment}
There is no gold-standard metric to evaluate the quality of times series data augmentations.
To assess the quality, we compared the augmented data from \emph{RATSpy} with the ones from \emph{tsaug}, 
% \hh{modified:}
using \gls{dtw} to measure the similarity between data points before and after augmentation.
For simple augmentations, such as Jittering, Repeat, and Drift, a high similarity score suggests that the augmentation quality is good and that the general structure of the input time series has remained intact, while for more complicated augmentations, such as amplitude and phase perturbation, where the inherent properties of the data are changed, 
as expected, very low degree of similarity can be observed.
Figure~\ref{fig:dtw_plots} highlights the \gls{dtw} score, showing $>80\%$ similarity for basic augmentations, such as Drift ($86.2\%$) and Reverse ($93.5\%$), as well as for augmentations where only a part of the series is changed, such as the random window time warp ($97.42\%$), whereas for \gls{app}, the similarity between the original and augmented time series drops to $53.6\%$.

\section{Conclusion} \label{conclusion}

In this report, we introduced \emph{\gls{rats}}, a Rust-based library for high-performance time series augmentation along with a Python wrapper, \emph{RATSpy}.
% \hh{slightly polished:}
\emph{\gls{rats}} comprises 17 basic, spectral (frequency-domain) and warping augmentation techniques and provides frequency-domain transforms
(\gls{dct} and \gls{fft}) in addition to a \gls{dtw} module for quantitative quality analysis. 
The design of \emph{\gls{rats}} using a uniform trait interface allows any combination of augmentation techniques to be chained into a pipeline and offers parallelisation using a single \texttt{parallel} flag.
Benchmarking results on 143 datasets clearly indicate that \emph{RATSpy} is faster and more memory-efficient than pure-Python
libraries, achieving up to 94.8\% faster running time and up to 47.9\% less peak memory usage compared to \emph{tsaug}.
% The running time of \emph{RATSpy} scales almost linearly with dataset size, and peak memory grows predictably with dataset size.
% \hh{this last statement sounds vague. precisely what are you trying to say?}\ws{I removed the statement. It is not really needed.}

%, confirming that the Rust core eliminates interpreter overhead and avoids redundant allocations.

\section*{Code Availability}
The \emph{\gls{rats}} library is open source and can be accessed at: \url{https://github.com/HyperVectors/RATS}

\section*{Acknowledgments}
This work has been partly funded by the Alexander von Humboldt Foundation through an Alexander von Humboldt Professorship awarded to Holger Hoos in 2022.
Part of the simulations and experiments were performed with computing resources granted by RWTH Aachen University.

% Further improvements that can be implemented in the future include extending the augmenters to handle multivariate time series data, and extending the per sample pipelining to pyFraug.

% \bibliography{bib}

\printbibliography[heading=bibintoc]

\clearpage
\appendix

\section{Detailed Results}\label{appsec:detailed_results}

% Font size macro so that table fits on page
\footnotesize

\begin{longtable}{l c c r r r r r r}
\caption{Detailed benchmarking results of \emph{RATSpy} versus \emph{tsaug} regarding both running time (\textbf{T}) and peak memory usage (\textbf{M}) across 143 datasets (all univariate equal-length datasets offered in aeon~\parencite{middlehurst_2024_aeon}).
Running time was measured in milliseconds (wall-clock time) and memory usage was measured in megabytes (MB).
Here, \textbf{N} denotes the number of samples in the dataset, \textbf{L} denotes the sequence length of the samples in the dataset, and $\mathbf{\Delta}$ denotes the percentage-wise improvement achieved by \emph{RATSpy} compared with \emph{tsaug} for the corresponding dataset.
Running time improvement is calculated as $\frac{t_{\text{tsaug}}-t_{\text{RATSpy}}}{t_{\text{tsaug}}} \times 100$ and 
peak memory usage improvement as $\frac{m_{\text{tsaug}}-m_{\text{RATSpy}}}{m_{\text{tsaug}}} \times 100$.}\\
\toprule
\multirow{2}{*}{\textbf{Dataset}} & \multirow{2}{*}{\textbf{N}} & \multirow{2}{*}{\textbf{L}} &
\multicolumn{2}{c}{\textbf{RATSpy}} &
\multicolumn{2}{c}{\textbf{tsaug}} &
\multicolumn{2}{c}{$\mathbf{\Delta} \, (\%)$} \\
\cmidrule(l{3pt}r{3pt}){4-5}\cmidrule(l{3pt}r{3pt}){6-7}
\cmidrule(l{3pt}r{3pt}){8-9}
  & & & \textbf{T} & \textbf{M} &
      \textbf{T} & \textbf{M} &
      \textbf{T} & \textbf{M} \\
\endfirsthead
\multicolumn{9}{c}%
{\tablename\ \thetable\ -- \textit{Continued from previous page}} \\    \midrule
\multirow{2}{*}{\textbf{Dataset}} & \multirow{2}{*}{\textbf{N}} & \multirow{2}{*}{\textbf{L}} &
\multicolumn{2}{c}{\textbf{RATSpy}} &
\multicolumn{2}{c}{\textbf{tsaug}} &
\multicolumn{2}{c}{$\mathbf{\Delta} \, (\%)$} \\
\cmidrule(l{3pt}r{3pt}){4-5}\cmidrule(l{3pt}r{3pt}){6-7}
\cmidrule(l{3pt}r{3pt}){8-9}
  & & & \textbf{T} & \textbf{M} &
      \textbf{T} & \textbf{M} &
      \textbf{T} & \textbf{M} \\ \midrule
\endhead
\hline \multicolumn{9}{r}{\textit{Continued on next page}} \\
\endfoot
\endlastfoot
\bottomrule
Adiac & 781 & 176 & 4.8 & 138.1 & 20 & 138.8 & 76.5 & 0.5 \\ 
ArrowHead & 211 & 251 & 3.4 & 129.9 & 7.8 & 129.9 & 56.8 & 0 \\ 
Beef & 60 & 470 & 1.1 & 133.4 & 4.7 & 133.4 & 76.8 & 0 \\ 
BeetleFly & 40 & 512 & 1 & 128.1 & 4 & 128.1 & 75.3 & 0 \\ 
BirdChicken & 40 & 512 & 0.9 & 125.7 & 4 & 125.7 & 76.7 & 0 \\ 
Car & 120 & 577 & 2.9 & 131.8 & 7.3 & 131.8 & 60.4 & 0 \\ 
CBF & 930 & 128 & 5.6 & 136.3 & 25 & 139.3 & 77.8 & 2.2 \\ 
ChlorineConcentration & 4307 & 166 & 50 & 167.1 & 128 & 200.4 & 60.6 & 16.6 \\ 
CinCECGTorso & 1420 & 1639 & 15 & 277.0 & 175 & 294.2 & 91.5 & 5.9 \\ 
Coffee & 56 & 286 & 0.9 & 128.1 & 3.9 & 128.1 & 76.2 & 0 \\ 
Computers & 500 & 720 & 3.3 & 161.8 & 27 & 161.8 & 87.8 & 0 \\ 
CricketX & 780 & 300 & 11 & 145.7 & 22 & 145.7 & 50.2 & 0 \\ 
CricketY & 780 & 300 & 4.4 & 149.7 & 24 & 149.7 & 81.7 & 0 \\ 
CricketZ & 780 & 300 & 10 & 147.9 & 22 & 147.9 & 51.1 & 0 \\ 
DiatomSizeReduction & 322 & 345 & 2.2 & 133.8 & 13 & 133.8 & 82.7 & 0 \\ 
DistalPhalanxOutlineAgeGroup & 539 & 80 & 5.2 & 134.7 & 11 & 134.7 & 55.6 & 0 \\ 
DistalPhalanxOutlineCorrect & 876 & 80 & 4.1 & 134.9 & 18 & 136.4 & 77.7 & 1.1 \\ 
DistalPhalanxTW & 539 & 80 & 5.4 & 132.4 & 11 & 133.9 & 54.2 & 1.1 \\ 
Earthquakes & 461 & 512 & 3.6 & 148.7 & 19 & 148.7 & 81.1 & 0 \\ 
ECG200 & 200 & 96 & 2.2 & 136.3 & 6.7 & 136.3 & 67.3 & 0 \\ 
ECG5000 & 5000 & 140 & 21 & 177.0 & 165 & 217.6 & 87.1 & 18.7 \\ 
ECGFiveDays & 884 & 136 & 10 & 136.4 & 19 & 138.7 & 45.3 & 1.6 \\ 
ElectricDevices & 16637 & 96 & 56 & 236.5 & 457 & 351.9 & 87.6 & 32.8 \\ 
FaceAll & 2250 & 131 & 24 & 151.4 & 56 & 156.0 & 56.3 & 2.9 \\ 
FaceFour & 112 & 350 & 1 & 132.9 & 6 & 132.9 & 83.5 & 0 \\ 
FacesUCR & 2250 & 131 & 24 & 151.0 & 52 & 154.9 & 53.8 & 2.5 \\ 
FiftyWords & 905 & 270 & 4.1 & 149.1 & 26 & 149.1 & 84.4 & 0 \\ 
Fish & 350 & 463 & 7.2 & 141.4 & 14 & 141.4 & 49.2 & 0 \\ 
FordA & 4921 & 500 & 28 & 295.8 & 248 & 360.3 & 88.5 & 17.9 \\ 
FordB & 4446 & 500 & 89 & 212.9 & 236 & 280.9 & 62.2 & 24.2 \\ 
GunPoint & 200 & 150 & 3.2 & 132.3 & 7.1 & 132.3 & 54.7 & 0 \\ 
Ham & 214 & 431 & 3.9 & 137.0 & 10 & 137.0 & 61.8 & 0 \\ 
HandOutlines & 1370 & 2709 & 20 & 363.3 & 278 & 465.7 & 92.8 & 22 \\ 
Haptics & 463 & 1092 & 15 & 176.0 & 32 & 176.0 & 52.8 & 0 \\ 
Herring & 128 & 512 & 1.1 & 136.3 & 7.7 & 136.3 & 85.2 & 0 \\ 
InlineSkate & 650 & 1882 & 34 & 184.5 & 89 & 222.2 & 61.1 & 17 \\ 
ItalyPowerDemand & 1096 & 24 & 3.1 & 131.8 & 19 & 135.6 & 84.1 & 2.8 \\ 
LargeKitchenAppliances & 750 & 720 & 18 & 174.9 & 39 & 174.9 & 53.8 & 0 \\ 
Lightning2 & 121 & 637 & 1 & 132.7 & 7.8 & 132.7 & 86.9 & 0 \\ 
Lightning7 & 143 & 319 & 2.5 & 134.9 & 6.6 & 134.9 & 61.3 & 0 \\ 
Mallat & 2400 & 1024 & 19 & 288.8 & 208 & 344.1 & 90.8 & 16.1 \\ 
Meat & 120 & 448 & 2.2 & 129.7 & 6.4 & 129.7 & 65.3 & 0 \\ 
MedicalImages & 1141 & 99 & 4.6 & 139.2 & 25 & 144.4 & 82 & 3.6 \\ 
MiddlePhalanxOutlineAgeGroup & 554 & 80 & 5.6 & 134.8 & 12 & 134.8 & 54.7 & 0 \\ 
MiddlePhalanxOutlineCorrect & 891 & 80 & 4 & 133.2 & 20 & 134.7 & 80.1 & 1.1 \\ 
MiddlePhalanxTW & 553 & 80 & 5.2 & 130.4 & 13 & 131.9 & 60.9 & 1.1 \\ 
MoteStrain & 1272 & 84 & 7.9 & 148.9 & 36 & 148.9 & 78.1 & 0 \\ 
NonInvasiveFetalECGThorax1 & 3765 & 750 & 101 & 250.3 & 236 & 270.0 & 57.1 & 7.3 \\ 
NonInvasiveFetalECGThorax2 & 3765 & 750 & 26 & 297.8 & 238 & 342.8 & 88.8 & 13.1 \\ 
OliveOil & 60 & 570 & 1.4 & 128.9 & 4.8 & 128.9 & 69.8 & 0 \\ 
OSULeaf & 442 & 427 & 3 & 143.7 & 17 & 143.7 & 82.9 & 0 \\ 
PhalangesOutlinesCorrect & 2658 & 80 & 26 & 152.8 & 54 & 161.3 & 51.3 & 5.3 \\ 
Phoneme & 2110 & 1024 & 17 & 263.8 & 185 & 329.2 & 90.6 & 19.8 \\ 
Plane & 210 & 144 & 2.4 & 127.4 & 6.7 & 128.2 & 63.9 & 0.6 \\ 
ProximalPhalanxOutlineAgeGroup & 605 & 80 & 5.5 & 126.9 & 12 & 129.1 & 57.2 & 1.7 \\ 
ProximalPhalanxOutlineCorrect & 891 & 80 & 9.2 & 135.5 & 18 & 137.0 & 48.8 & 1.1 \\ 
ProximalPhalanxTW & 605 & 80 & 3.3 & 133.4 & 13 & 133.4 & 75.4 & 0 \\ 
RefrigerationDevices & 750 & 720 & 19 & 174.4 & 38 & 174.4 & 49.2 & 0 \\ 
ScreenType & 750 & 720 & 5.1 & 178.5 & 39 & 178.5 & 87.1 & 0 \\ 
ShapeletSim & 200 & 500 & 3.9 & 134.0 & 9.4 & 134.0 & 58.5 & 0 \\ 
ShapesAll & 1200 & 512 & 8.7 & 191.2 & 50 & 191.2 & 82.8 & 0 \\ 
SmallKitchenAppliances & 750 & 720 & 18 & 175.1 & 39 & 175.1 & 53.6 & 0 \\ 
SonyAIBORobotSurface1 & 621 & 70 & 6.9 & 245.2 & 13 & 245.2 & 47.2 & 0 \\ 
SonyAIBORobotSurface2 & 980 & 65 & 8.9 & 144.6 & 19 & 144.6 & 54.2 & 0 \\ 
StarLightCurves & 9236 & 1024 & 61 & 634.8 & 862 & 892.4 & 92.9 & 28.9 \\ 
Strawberry & 983 & 235 & 12 & 145.5 & 25 & 145.5 & 50.6 & 0 \\ 
SwedishLeaf & 1125 & 128 & 4.4 & 136.2 & 30 & 139.9 & 85.7 & 2.7 \\ 
Symbols & 1020 & 398 & 17 & 146.6 & 35 & 158.7 & 50.8 & 7.6 \\ 
SyntheticControl & 600 & 60 & 3 & 137.2 & 13 & 137.2 & 77.5 & 0 \\ 
ToeSegmentation1 & 268 & 277 & 3.9 & 132.6 & 9.4 & 132.6 & 59.2 & 0 \\ 
ToeSegmentation2 & 166 & 343 & 3 & 128.4 & 10 & 128.4 & 69.7 & 0 \\ 
Trace & 200 & 275 & 3.3 & 129.6 & 7.6 & 129.6 & 56.5 & 0 \\ 
TwoLeadECG & 1162 & 82 & 4.4 & 147.8 & 24 & 147.8 & 82.1 & 0 \\ 
TwoPatterns & 5000 & 128 & 53 & 187.5 & 149 & 209.5 & 64.3 & 10.5 \\ 
UWaveGestureLibraryAll & 4478 & 945 & 30 & 412.0 & 385 & 508.9 & 92.2 & 19.1 \\ 
UWaveGestureLibraryX & 4478 & 315 & 70 & 216.8 & 179 & 248.3 & 60.6 & 12.7 \\ 
UWaveGestureLibraryY & 4478 & 315 & 21 & 246.9 & 174 & 258.5 & 87.5 & 4.5 \\ 
UWaveGestureLibraryZ & 4478 & 315 & 68 & 212.2 & 164 & 244.6 & 58.3 & 13.3 \\ 
Wafer & 7164 & 152 & 30 & 200.2 & 217 & 252.7 & 85.7 & 20.8 \\ 
Wine & 111 & 234 & 1.5 & 126.7 & 5 & 126.7 & 69.2 & 0 \\ 
WordSynonyms & 905 & 270 & 4.7 & 151.5 & 27 & 151.5 & 82.6 & 0 \\ 
Worms & 258 & 900 & 7.1 & 147.1 & 16 & 147.1 & 55.8 & 0 \\ 
WormsTwoClass & 258 & 900 & 3.1 & 144.4 & 17 & 144.4 & 82.3 & 0 \\ 
Yoga & 3300 & 426 & 59 & 211.4 & 143 & 254.4 & 58.2 & 16.9 \\ 
ACSF1 & 200 & 1460 & 2.2 & 154.9 & 18 & 154.9 & 88.5 & 0 \\ 
AllGestureWiimoteX & 1000 & 0 & 21 & 175.2 & 40 & 175.2 & 47.2 & 0 \\ 
AllGestureWiimoteY & 1000 & 0 & 5.4 & 174.9 & 40 & 174.9 & 86.7 & 0 \\ 
AllGestureWiimoteZ & 1000 & 0 & 20 & 175.2 & 40 & 175.2 & 48.7 & 0 \\ 
BME & 180 & 128 & 1.2 & 124.6 & 7 & 124.6 & 82.9 & 0 \\ 
EthanolLevel & 1004 & 1751 & 9.7 & 233.0 & 128 & 287.0 & 92.5 & 18.8 \\ 
FreezerRegularTrain & 3000 & 301 & 15 & 180.1 & 93 & 207.2 & 83.1 & 13.1 \\ 
FreezerSmallTrain & 2878 & 301 & 15 & 170.1 & 89 & 197.5 & 82.7 & 13.9 \\ 
GunPointAgeSpan & 451 & 150 & 2.5 & 131.0 & 12 & 132.5 & 79.4 & 1.1 \\ 
GunPointMaleVersusFemale & 451 & 150 & 2.4 & 130.5 & 10 & 132.0 & 77.9 & 1.1 \\ 
GunPointOldVersusYoung & 451 & 150 & 2.2 & 128.0 & 12 & 129.5 & 82 & 1.2 \\ 
InsectEPGRegularTrain & 311 & 601 & 5.9 & 152.3 & 13 & 152.3 & 54.7 & 0 \\ 
InsectEPGSmallTrain & 266 & 601 & 2.2 & 138.2 & 12 & 138.2 & 82.7 & 0 \\ 
PickupGestureWiimoteZ & 100 & 0 & 1.1 & 126.6 & 4.7 & 126.6 & 76.2 & 0 \\ 
PigAirwayPressure & 312 & 2000 & 3.2 & 163.1 & 42 & 176.9 & 92.6 & 7.8 \\ 
PigArtPressure & 312 & 2000 & 3.2 & 161.2 & 38 & 170.9 & 91.8 & 5.7 \\ 
PigCVP & 312 & 2000 & 3.7 & 162.1 & 39 & 176.1 & 90.7 & 8 \\ 
PowerCons & 360 & 144 & 1.7 & 125.4 & 9.7 & 126.9 & 82.3 & 1.2 \\ 
ShakeGestureWiimoteZ & 100 & 0 & 1.5 & 128.9 & 4.5 & 128.9 & 67.8 & 0 \\ 
SmoothSubspace & 300 & 15 & 2 & 133.1 & 8.1 & 133.1 & 75.7 & 0 \\ 
UMD & 180 & 150 & 1.4 & 126.5 & 6.2 & 126.5 & 76.9 & 0 \\ 
Fungi & 204 & 201 & 1.6 & 125.8 & 7.4 & 125.8 & 78.4 & 0 \\ 
GesturePebbleZ1 & 304 & 0 & 2.3 & 135.5 & 13 & 135.5 & 82.5 & 0 \\ 
GesturePebbleZ2 & 304 & 0 & 2.5 & 135.4 & 13 & 135.4 & 80.7 & 0 \\ 
HouseTwenty & 135 & 3000 & 1.8 & 152.0 & 17 & 152.0 & 89.7 & 0 \\ 
DodgerLoopDay & 158 & 288 & 1.6 & 126.3 & 6.3 & 126.3 & 75 & 0 \\ 
DodgerLoopWeekend & 158 & 288 & 2.4 & 129.0 & 6.2 & 129.0 & 61.2 & 0 \\ 
DodgerLoopGame & 158 & 288 & 1.7 & 127.1 & 6.1 & 127.1 & 71.7 & 0 \\ 
SemgHandGenderCh2 & 900 & 1500 & 9.2 & 201.9 & 100 & 243.8 & 90.9 & 17.2 \\ 
SemgHandMovementCh2 & 900 & 1500 & 8.7 & 197.5 & 97 & 239.3 & 91 & 17.4 \\ 
SemgHandSubjectCh2 & 900 & 1500 & 8.8 & 201.7 & 94 & 243.9 & 90.7 & 17.3 \\ 
MixedShapesSmallTrain & 2525 & 1024 & 19 & 267.1 & 216 & 289.6 & 91.1 & 7.8 \\ 
EOGHorizontalSignal & 724 & 1250 & 6.4 & 184.7 & 70 & 212.2 & 90.9 & 13 \\ 
EOGVerticalSignal & 724 & 1250 & 27 & 159.9 & 72 & 188.0 & 62.5 & 14.9 \\ 
GestureMidAirD1 & 338 & 360 & 2.2 & 134.2 & 12 & 134.2 & 81.9 & 0 \\ 
GestureMidAirD2 & 338 & 360 & 3 & 136.2 & 12 & 136.2 & 76.3 & 0 \\ 
GestureMidAirD3 & 338 & 360 & 2.2 & 135.9 & 11 & 135.9 & 80.3 & 0 \\ 
Rock & 70 & 2844 & 1.3 & 151.4 & 13 & 151.4 & 90.2 & 0 \\ 
Crop & 24000 & 46 & 89 & 225.0 & 677 & 412.8 & 86.8 & 45.5 \\ 
Chinatown & 365 & 24 & 2.2 & 126.1 & 7.6 & 126.8 & 71 & 0.6 \\ 
MelbournePedestrian & 3633 & 24 & 8.7 & 143.1 & 73 & 171.8 & 88.1 & 16.7 \\ 
AsphaltObstacles & 781 & 0 & 17 & 158.7 & 47 & 171.6 & 62.8 & 7.5 \\ 
AsphaltPavementType & 2111 & 0 & 23 & 428.8 & 338 & 560.6 & 93.1 & 23.5 \\ 
AsphaltRegularity & 1502 & 0 & 45 & 404.0 & 416 & 554.2 & 89 & 27.1 \\ 
Colposcopy & 200 & 180 & 1.4 & 126.1 & 6 & 126.1 & 76.3 & 0 \\ 
RightWhaleCalls & 12896 & 4000 & 204 & 2406.8 & 3914 & 3993.2 & 94.8 & 39.7 \\ 
SharePriceIncrease & 1930 & 60 & 7.7 & 141.5 & 34 & 147.4 & 77.5 & 4 \\ 
CatsDogs & 275 & 14773 & 19 & 433.2 & 301 & 544.2 & 93.4 & 20.4 \\ 
BinaryHeartbeat & 409 & 18530 & 157 & 349.5 & 464 & 639.0 & 66.1 & 45.3 \\ 
DucksAndGeese & 100 & 236784 & 78 & 2495.2 & 1486 & 3141.0 & 94.8 & 20.6 \\ 
InsectSound & 50000 & 600 & 265 & 1630.5 & 3365 & 2531.0 & 92.1 & 35.6 \\ 
AbnormalHeartbeat & 606 & 3053 & 32 & 535.9 & 458 & 710.1 & 93 & 24.5 \\ 
ElectricDeviceDetection & 4390 & 256 & 52 & 195.2 & 148 & 228.5 & 64.8 & 14.6 \\ 
KeplerLightCurves & 1319 & 4767 & 24 & 407.3 & 416 & 554.6 & 94.2 & 26.5 \\ 
Sleep & 569100 & 178 & 2222 & 4825.0 & 21557 & 9263.6 & 89.7 & 47.9 \\ 
FaultDetectionA & 13640 & 5120 & 1563 & 2854.0 & 4975 & 4986.0 & 68.6 & 42.8 \\ 
FaultDetectionB & 13640 & 5120 & 318 & 2790.4 & 4841 & 4954.6 & 93.4 & 43.7 \\ 
NerveDamage & 204 & 1500 & 2.8 & 152.9 & 17 & 152.9 & 83.9 & 0 \\ 
Epilepsy2 & 11500 & 178 & 54 & 270.7 & 360 & 315.6 & 85 & 14.2 \\
\hline
Average & & & & & & & 74.5 & 7.3 \\
\bottomrule
\label{tab:pipeline_benchmarks_all}
\end{longtable}
\normalsize

\end{document}